\documentclass[a4paper,fleqn]{cas-sc}

\usepackage[round,sort]{natbib}

\usepackage{float}
\restylefloat{figure}
\usepackage{cleveref}
\usepackage{multirow}
\usepackage{rotating}
\usepackage{soul}
\usepackage{longtable}
\usepackage{ltablex}
\usepackage{caption, booktabs}

\def\tsc#1{\csdef{#1}{\textsc{\lowercase{#1}}\xspace}}
\tsc{WGM}
\tsc{QE}
\tsc{EP}
\tsc{PMS}
\tsc{BEC}
\tsc{DE}

\begin{document}
\let\WriteBookmarks\relax
\def\floatpagepagefraction{1}
\def\textpagefraction{.001}
\shorttitle{}
 \shortauthors{Khadivi et al.}

\title [mode = title]{Deep reinforcement learning for machine scheduling: Methodology, the state-of-the-art, and future directions}                      

\author[1]{Maziyar Khadivi}[type=editor, auid=000,bioid=1,]
\ead{mazy1996@uvic.ca}
\credit{Conceptualization, Formal analysis, Investigation, Data Curation, Visualization, Writing - Original Draft, Writing - Review \& Editing, Supervision, Project administration}

\author[2]{Todd Charter}[type=editor, orcid=0000-0001-5982-255X, auid=000,bioid=1,]
\ead{toddch@uvic.ca}
\credit{Investigation, Visualization, Writing - Original Draft, Writing - Review \& Editing}

\author[1]{Marjan Yaghoubi}[type=editor, auid=000,bioid=1,]
\ead{marjanyaghoubi@uvic.ca}
\credit{Writing - Original Draft}

\author[1]{Masoud Jalayer}[type=editor, orcid=0000-0001-8013-8613, auid=000,bioid=1,]
\ead{masoudjalayer@uvic.ca}
\credit{Writing - Original Draft}

\author[2]{Maryam Ahang}[type=editor, auid=000,bioid=1,]
\ead{maryamahang@uvic.ca}
\credit{Writing - Original Draft}

\author[2]{Ardeshir Shojaeinasab}[type=editor, auid=000,bioid=1,]
\ead{ardeshir@uvic.ca}
\credit{Writing - Original Draft}

\author[1,2]{Homayoun Najjaran}[type=editor, orcid=0000-0002-3550-225X, auid=000,bioid=1]
\cormark[1]
\ead{najjaran@uvic.ca}
\credit{Funding acquisition, Writing - Review \& Editing, Supervision} 

\address[1]{Department of Mechanical Engineering, University of Victoria, Victoria BC, V8P 5C2, Canada}

\address[2]{Department of Electrical and Computer Engineering, University of Victoria, Victoria BC, V8P 5C2, Canada}

\cortext[cor1]{Corresponding author}

\begin{abstract}
Machine scheduling aims to optimally assign jobs to a single or a group of machines while meeting manufacturing rules as well as job specifications. Optimizing the machine schedules lead to significant reduction in operational costs, adherence to customer demand, and rise in production efficiency. Despite its benefits for the industry, machine scheduling remains a challenging combinatorial optimization problem to be solved, inherently due to its Non-deterministic Polynomial-time (NP) hard nature. Deep Reinforcement Learning (DRL) has been regarded as a foundation for \emph{"artificial general intelligence"} with promising results in tasks such as gaming and robotics. Researchers have also aimed to leverage the application of DRL, attributed to extraction of knowledge from data, across variety of machine scheduling problems since 1995. This paper presents a comprehensive review and comparison of the methodology, application, and the advantages and limitations of different DRL-based approaches. Further, the study categorizes the DRL methods based on the computational components including conventional neural networks, encoder-decoder architectures, graph neural networks and metaheuristic algorithms. Our literature review concludes that the DRL-based approaches surpass the performance of exact solvers, heuristics, and tabular reinforcement learning algorithms in computation speed and generating near-global optimal solutions. They have been applied to static or dynamic scheduling of different machine environments with different job characteristics. Nonetheless, the existing DRL-based schedulers face limitations not only in considering complex operational constraints, and configurable multi-objective optimization but also in dealing with generalization, scalability, intepretability, and robustness. Therefore, addressing these challenges shapes future work in this field. This paper serves the researchers to establish a proper investigation of sate of the art and research gaps in DRL-based machine scheduling and can help the experts and practitioners choose the proper approach to implement DRL for production scheduling.

\end{abstract}



\maketitle
\section{Introduction} \label{sec: introduction}

\begin{table}[]
\caption{Table of abbreviations}
\label{tab:abbreviations}
\centering
\begin{tabular}{llll}

\toprule
A2C   & Advantage Actor-Critic                   & JSSP    & Job Shop Scheduling Problem              \\
A3C   & Asynchronous Advantage Actor-Critic      & KPI     & Key Performance Indicator                \\
AC    & Actor-Critic                             & L2C     & Learn to Construct                       \\
ANN   & Artificial Neural Network                & L2I     & Learn to Improve                         \\
CNN   & Convolutional Neural Network             & LSTM    & Long Short Term Memory                   \\
DAG   & Directed Acyclic Graph                   & MA      & Multi Agent                              \\
DDPG  & Deep Deterministic Policy Gradient       & MADRL   & Multi-Agent Deep Reinforcement Learning  \\
DDPG  & Deep Deterministic Policy Gradient       & MDP     & Markov Decision Process                  \\
DNN   & Deep Neural Network                      & MILP    & Mixed-Integer Linear Programming         \\
DQN   & Deep Q Network                           & MPNN    & Message Passing Neural Network           \\
DRL   & Deep Reinforcement Learning              & NP      & Non-deterministic Polynomial-time        \\
EDA   & Estimation of Distribution Algorithm     & PFSSP   & Permutation Flow Shop Scheduling Problem \\
FFSSP & Flexible Flow Shop Scheduling Problem    & PG      & Policy Gradient                          \\
FJSSP & Flexible Job Shop Scheduling Problem     & PN      & Pointer Network                          \\
FNN   & Feedforward Neural Network               & PPO     & Proximal Policy Optimisation             \\
FSSP  & Flow Shop Scheduling Problem             & ReLU    & Rectified Linear Unit                    \\
GA    & Genetic Algorithm                        & RL      & Reinforcement Learning                   \\
GAT   & Graph Attention Network                  & RNN     & Recurrent Neural Network                 \\
GCN   & Graph Convolutional Network              & SA      & Single Agent                             \\
GIN   & Graph Isomorphism Network                & SARSA   & State–Action–Reward–State–Action         \\
GNN   & Graph Neural Network                     & Seq2Seq & Sequence-to-Sequence                     \\
GP    & Genetic Programming                      & TD      & Temporal-Difference                      \\
GRU   & Gated Recurrent Unit                     & TDNN    & Time Delay Neural Network                \\
HDRL  & Hierarchical Deep Reinforcement Learning & TRPO    & Trust Region Policy Optimization         \\
IG    & Iterated Greedy                          &         &                                         \\
\bottomrule

\end{tabular}
\end{table}

Nowadays, manufacturing companies must deal with shortening production lead times, mass customization and lowering  production cost to remain competitive in highly uncertain market conditions and to satisfy customer demand in the shortest possible time \citep{Panzer2021}. Scheduling, as a decision-making process, plays a key role in helping manufacturers cope with these challenges. It is performed in most production and manufacturing systems on a regular basis to determine the allocation of available resources to the tasks over given time periods considering optimization of single or multiple objectives. The resources are machines in a workshop, while tasks are operations of jobs in a production process. To schedule the operations on the machines, different objectives can be considered such as minimization of the number of tasks completed after their deadline or minimization of the completion time of the last task \citep{pinedo1992scheduling}.\\

Since firms most often require to quickly generate a schedule for processing a large number of jobs every day, even a slight improvement in the scheduler can result in significant gains \citep{Li2022}. Machine scheduling problems have computational complexity of NP-hard posing a big challenge for algorithms to find an efficient solution in polynomial time \citep{mazyavkina2021reinforcement}. A scheduling algorithm is considered good when it can meet the operational constraints, have computational efficiency, and achieve solution quality \citep{Li2022}. Exact solvers, such as the branch and bound and cutting-plane methods, search the solution space based on enumeration to find the global optimum \citep{dong2022minimizing}. For this reason, they exhibit good performance only when they are applied to small-scale problems \citep{wolsey2020integer}. On the other hand, the exact solvers suffer from the curse of dimensionality in more involved problems, making the runtime exhaustively long \citep{dong2022minimizing}. To deal with medium- and large-scale problems which is often the case in practice, approximation methods including heuristics and metaheuristics have been proposed as they offer a balance between the solution quality and the computation time \citep{dong2022minimizing}. Heuristics, priority rules, or dispatching rules are often used as synonyms for machine scheduling. They basically assign a value to each waiting job according to a rule, and then the job with the minimum value is selected for the execution \citep{panwalkar1977survey}. Even though heuristics can generate a solution in a short time, their solution quality is low \citep{dong2022minimizing}. Metaheuristics such as the particle swarm optimization algorithm, tabu search algorithm, and genetic algorithms are mainly nature-inspired algorithms that are problem independent \citep{rahman2021nature}. Unlike heuristics, they can approximate a near optimal solution even for the large-scale problems but at the cost of slow convergence speed which makes them practically infeasible for industrial implementations \citep{dong2022minimizing}. Deep reinforcement learning (DRL) has emerged as an alternative approach to combat the limitations of the traditional machine scheduling algorithms. DRL leverages the advantages of both deep learning and reinforcement learning to deliver a near optimal solution in a short computation time .\\
Reinforcement learning works based on Markov Decision Process (MDP) and consists of an agent and an environment that interact with each other dynamically \citep{sutton2018reinforcement}. The agent changes the state of the environment by taking specific actions and the environment returns a reward to the agent as feedback to the performed action. The agent iteratively interacts with the environment and gradually learns to receive more positive rewards by taking better actions when faced with particular states in the environment. The function that guides the agent to take a particular action at each state is called the policy \citep{sutton2018reinforcement}. Tabular RL approaches discretize the state and action space using a lookup table (also called Q-table). Use of a lookup table causes two key issues. First, the  RL agent will not be able to return Q-values for previously unseen states \citep{witty2021measuring}. Second, the number of states and actions grows exponentially in the case of high-dimensional problems \citep{bellman2015applied}. These issues lead to low learning efficiency, intensive requirement of memory, and degraded performance \citep{lange2012autonomous}. To cope with the limits of tabular RL methods, deep learning and RL can be fused together to leverage the generalization power of deep neural networks in large and complex problems. Deep learning is a subfield of machine learning that is usually used for supervised and unsupervised learning \citep{wang2021deep}. Deep learning models in their basic form are composed of one or more hidden layers of neurons that connect an input layer to an output layer \citep{wang2021deep}. They are well suited for processing natural signals such as voice, text, and images for which the true data distribution is unknown \textit{a priori}. The weights of neurons residing in the hidden layers are optimized using their derivatives such that a loss function is minimized \citep{Bengio2021}. Deep learning models have excelled in numerous tasks including image classification, pattern recognition, natural language processing, and recommender systems \citep{Bengio2021}. The basic and widely used models in deep learning are Feedforward Neural Networks (FNN), Recurrent Neural Netowrks (RNN), and Convolutional Neural Networks (CNN). The more advanced neural network architectures are built by a combination of the above-mentioned networks \citep{goodfellow2016deep}.

\subsection{Motivation} \label{sec: motivation}
\begin{figure}[H]
    \centering
    \includegraphics[width=15cm]{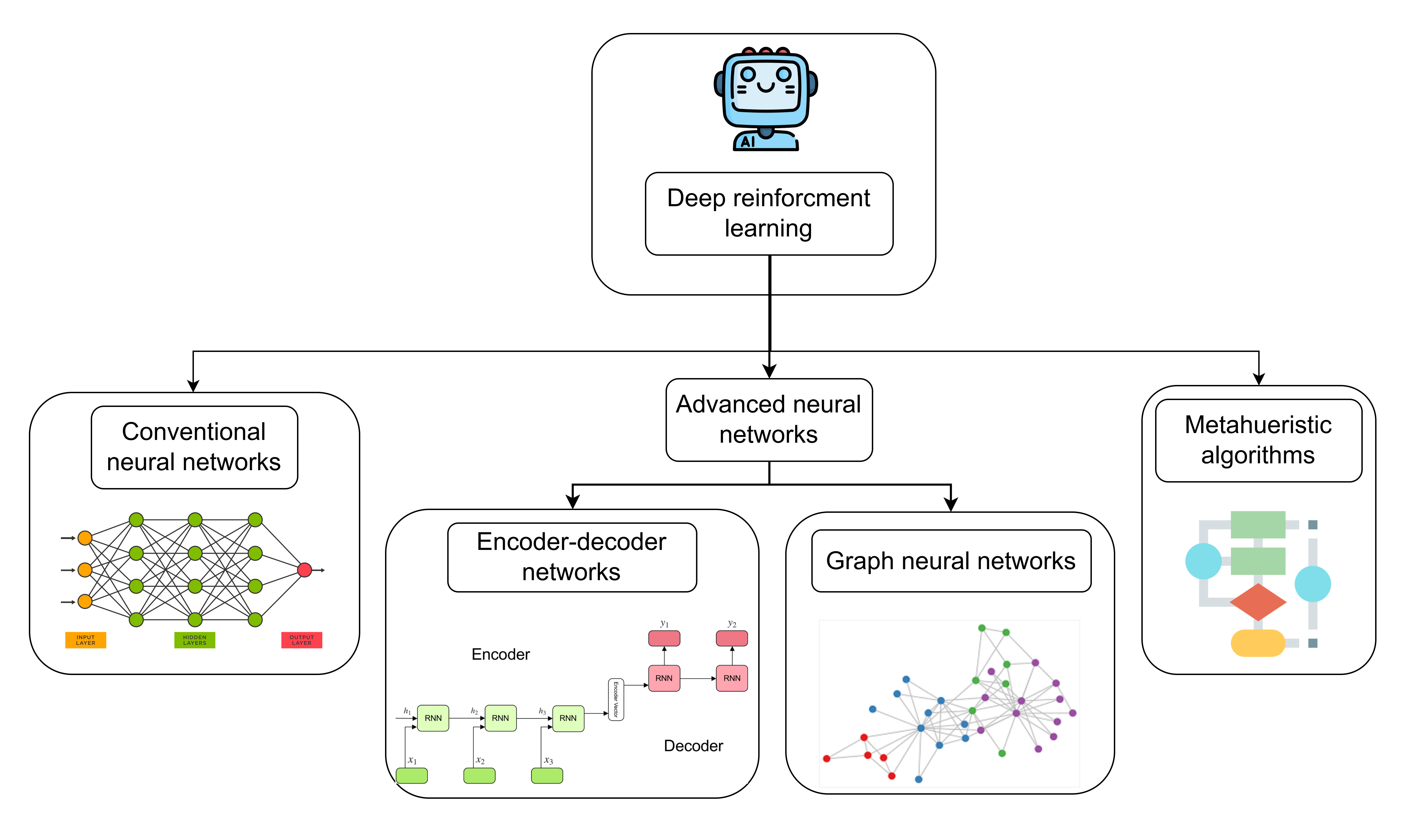}
    \caption{Classification of deep reinforcement learning approaches to optimize machine scheduling problems according to their computational component}
    \label{fig:approaches}
\end{figure}

Our literature review on the application of DRL in machine scheduling indicates that DRL models have been adopted to solve machine scheduling problems with four different computational components. In the most basic form, conventional neural networks, including FNN, RNN, and CNN are used as the function approximator in the DRL algorithm. In this paper, we refer to them as \textbf{conventional DRL} models. The second and third ideas are inspired by advanced deep learning architectures, i.e., encoder-decoder architectures and Graph Neural Networks (GNN), respectively. They achieve outstanding performance in terms of both scalability and generalization when applied to combinatorial optimization problems, herein machine scheduling problems. Encoder-decoder architectures and GNNs were developed originally for rearranging the sequence of words in machine translation tasks and processing graph-structured data, respectively \citep{vesselinova2020learning}. Since the scheduling problem instances can be formulated as a sequential data or a graph-structured data depending on the machine environment, they can be fed as data points to the encoder-decoder architectures or GNNs to learn latent, low-dimensional representation of the problem instances \citep{vesselinova2020learning}. The learnable parameters of the above architectures are optimized through DRL \citep{bello2016neural,kool2018attention}. In other words, the neural networks of these structures act as the function approximator of the RL agent \citep{bello2016neural,kool2018attention}. After the feature vector representations of the problem instances are learnt, they can be used to increase the probability of generating the desired sequence of processing the jobs on the machines \citep{Liang2022}. Since encoder-decoders and GNNs are among the advanced neural networks, we refer to the DRL models that are using them as \textbf{advanced DRL} models. The last approach uses a DRL model to make decisions alongside an optimization algorithm, interactively \citep{Bengio2021}. A master algorithm such as a metaheuristic controls the higher level decisions while repeatedly calling a DRL agent throughout its execution to assist in lower lever decisions \citep{Bengio2021}. We use a similar terminology that is first introduced in \citep{Bengio2021} to refer this type of algorithmic structure, i.e., \textbf{metaheuristic-based DRL}. \Cref{fig:approaches} illustrates the classification of DRL approaches according to their computational component.

\subsection{Related work} \label{sec: relatedwork}
The review papers and surveys that are related to the present study can be categorized based on the methodology and application domain that they considered within their scope of work. \Cref{tab:surveys} summarizes the scope of the previously reported surveys and review papers. As depicted, the majority of review papers surveyed the application of RL in production systems and combinatorial optimization problems. The papers that reviewed the application of RL in production systems, mainly surveyed tabular RLs and conventional DRLs. Their scope of review covered production-related topics, including capacity planning, purchase and supply management, process control, facility resource planning, inventory management, and production scheduling. They reviewed the studies related to production scheduling as one category and different machine environments (see \Cref{sec: machine_env}) were not discussed separately. It is insightful to review DRL production scheduling problems based on their machine environment since each production environment inherits unique properties and level of complexity; therefore, they demand different algorithmic solutions for scheduling. The papers that surveyed the application of RL in combinatorial optimization problems (where machine scheduling appears as a subfield), only covered reviewing advanced DRL methods. They provided a general overview of how DRL can be used to train encoder-decoder architectures and GNNs to solve combinatorial optimization problems. However, they did not review the studies that applied these methods particularly on machine scheduling problems. There is only one study which focused on reviewing the machine scheduling using RL but their scope of review includes only tabular RL and conventional DRL. This paper provides an inclusive review of all DRL methods applied to solve machine scheduling problems. More precisely, our literature review is complementary to the existing ones as we provide the essential background and classification of the DRL methods, review their application to machine scheduling, and discuss the benefits and limitations of different DRL methods with respect to different machine environments.

\begin{table}[ht]
\caption{Surveys and reviews of DRL in production systems, combinatorial optimization, and machine scheduling}
\label{tab:surveys}
\resizebox{\columnwidth}{!}{
\centering
\begin{tabular}{@{\extracolsep{4pt}}m{3.5 cm}m{1.5 cm}m{2.25 cm}m{2.25 cm}m{2.25 cm}m{2 cm}m{2 cm}m{2 cm}}

\toprule
\textbf{Authors (Year)} & \multicolumn{4}{c}{\textbf{Methodology}} & \multicolumn{3}{c}{\textbf{Application domain}} \\ \cline{2-5} \cline{6-8}

 & Tabular RL & Conventional DRL & Advanced DRL & Metaheuristic-based DRL & Production systems & Combinatorial Optimization & Machine scheduling \\

\midrule
\citet{priore2014dynamic} & \checkmark & - & - & - & \checkmark & - & -\\

\citet{vesselinova2020learning} & - & - & \checkmark & - & - & \checkmark & -\\ 

\citet{Panzer2021} & - & \checkmark & - & - & \checkmark & - & -\\ 

\citet{kayhan2021reinforcement} & \checkmark & \checkmark & - & - & - & - & \checkmark \\

\citet{waubert2022reliability} & - & \checkmark & - & - & \checkmark & - & - \\

\citet{Kotary2021} & - & - & \checkmark & - & - & \checkmark & - \\

\citet{mazyavkina2021reinforcement} & - & - & \checkmark & - & - & \checkmark & - \\

\citet{Bengio2021} & - & - & \checkmark & \checkmark & - & \checkmark & - \\

\citet{cappart2021combinatorial} & - & - & \checkmark & - & - & \checkmark & - \\

\citet{esteso2022reinforcement} & \checkmark & \checkmark & - & - & \checkmark & - & - \\ \hline

This study  & - & \checkmark & \checkmark & \checkmark & - & - & \checkmark \\

\bottomrule
\end{tabular}}
\end{table}

\subsection{Goal} \label{sec: goal}
This review focuses on the DRL methodologies and their application in machine scheduling based on the techniques presented in \Cref{fig:approaches}. We first survey all DRL methods applied to machine scheduling problems and then comprehensively review the main contributions of the previous works that used DRL to schedule different machine environments. Thus, we have expanded on the previously published surveys by including all DRL methods and also presenting their application in different machine scheduling environments. It provides an overview of ongoing research in DRL for machine scheduling to serve scholars in identifying research gaps and future research directions. This review also assists industry practitioners in considering possible implementation scenarios and motivates them to deploy research findings in production and manufacturing systems. To further motivate the goal of the present work, we define the following Research Questions (RQ) and seek to find answer for them with the aid of the present literature review:

\begin{enumerate}
    \item RQ1. What are the core DRL approaches employed to solve machine scheduling problems?
    \item RQ2. What is their applicability to different machine environments?
    \item RQ3. What are the benefits and limitations of each approach?
    \item RQ4. What are the current trends and existing challenges of DRL in machine scheduling?
    \item RQ5. What future research directions can address existing challenges of DRL application in machine scheduling?
\end{enumerate}

\subsection{Contribution}
So far we have introduced the algorithmic challenges in machine scheduling in \Cref{sec: introduction}, the motivation behind using DRL in \Cref{sec: motivation}, the related works in \Cref{sec: relatedwork} and the goal of our literature review in \Cref{sec: goal}. We contribute to the literature by answering the aforementioned research questions through the remainder of this paper. \Cref{sec:Preliminaries} defines the basics of machine scheduling essential to fully understand the content of the paper. \Cref{sec:method} describes the minimal prerequisites of Markov decision process, DRL, encoder-decoder architectures, and GNNs. \Cref{sec: approaches} answers RQ1 by explaining different DRL approaches depicted in \Cref{fig:approaches}. \Cref{sec:applications} answers RQ2 by a comprehensive review of the papers that applied DRL to machine scheduling problems. \Cref{sec:discussion} provides details for RQ3 and RQ4 by analysing the trend in using different DRL approaches, comparing the advantages and limits of each approach, and highlighting the existing challenges. RQ5 is answered in \Cref{sec:future directions} where the future avenues for research are discussed. Finally, \Cref{sec: conclusions} summarizes the advantages, disadvantages and limitations of using DRL for machine scheduling discussed in the surveyed literature. This paper requires the use of abbreviations, which are presented in \Cref{tab:abbreviations}.

\section{Machine scheduling} \label{sec:Preliminaries}
Machine scheduling deals with the sequencing of jobs to be processed by machines so that an objective function is optimized under certain constraints. The objective function often relates to processing jobs in a timely manner, reducing production costs, and maximizing machine utilization. Generally in machine scheduling environments, each machine is restricted to processing only one job at a time and each job may not be processed on more than a single machine simultaneously \citep{graham1979optimization}.

\subsection{Scheduling function in an enterprise}
\begin{figure}[H]
    \centering
    \includegraphics[width=10cm]{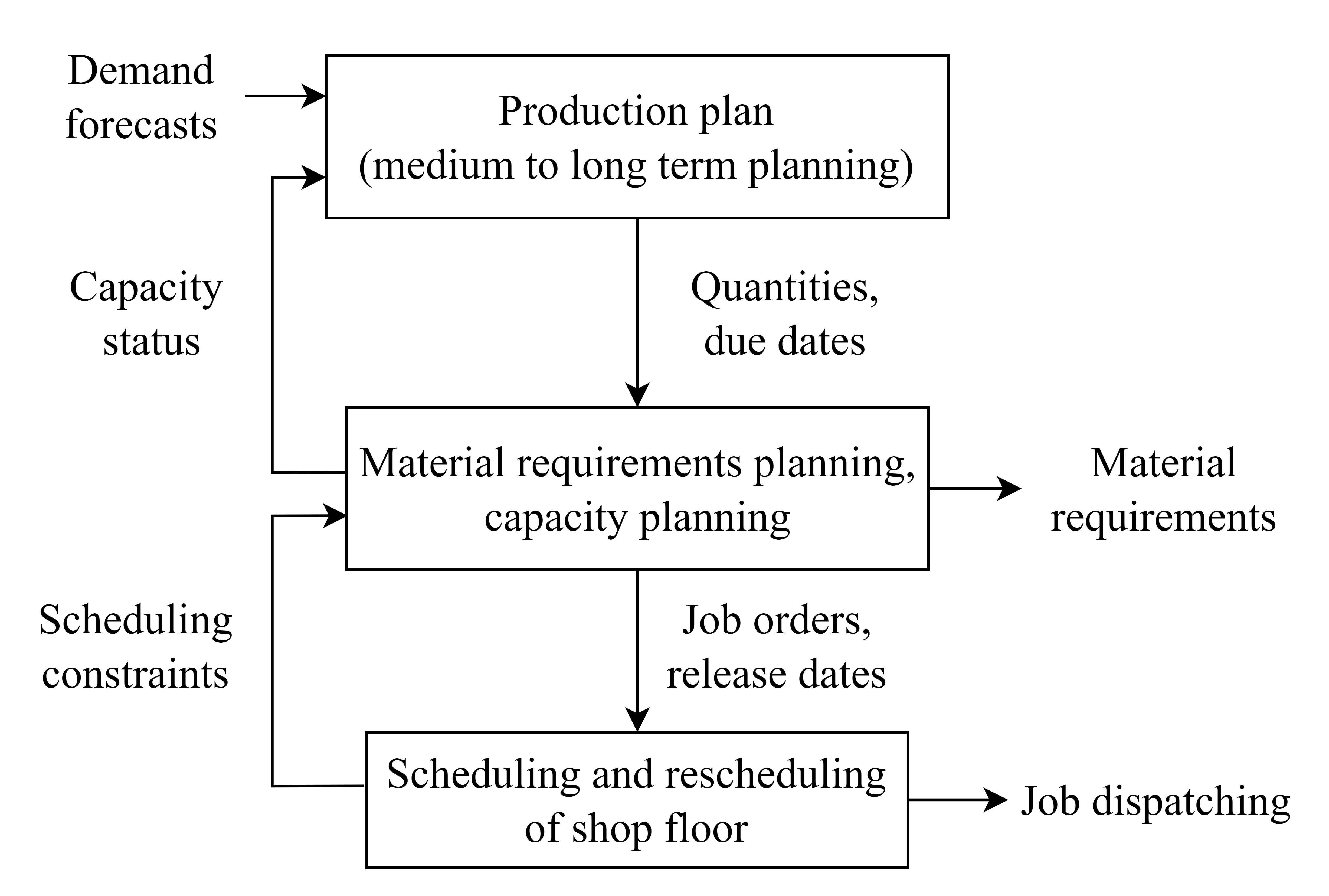}
    \caption{Information flow between scheduling function and other modules in an enterprise (adopted from \citep{pinedo1992scheduling})}
    \label{fig:ERP}
\end{figure}

Scheduling is among the main functions of manufacturing execution system that is often in interaction with many other functions at the enterprise level \citep{shojaeinasab2022intelligent}. Through enterprise resource planning modules, the scheduling system can give all departments at the enterprise level access to the scheduling information. In turn, it can receive up-to-date information about the statuses of jobs and machines. The input to the scheduling process is affected by the production plan that optimizes the firm’s medium- to long-term resource allocation and overall product mix based on demand forecasts, inventory levels, and available capacity of resources. Another decision making module that scheduling closely interacts with is material requirements planning. Since a job order can be scheduled to be processed only when its required resources and raw materials are available at the specified time periods, the material requirements planning system and the scheduling system jointly make decision about the release dates of all jobs \citep{pinedo1992scheduling}. \Cref{fig:ERP} depicts a diagram of information flow between scheduling function and other functions in an enterprise.

\subsection{Machine environment} \label{sec: machine_env}
\begin{figure} [H]
    \centering
    \includegraphics[width=10cm]{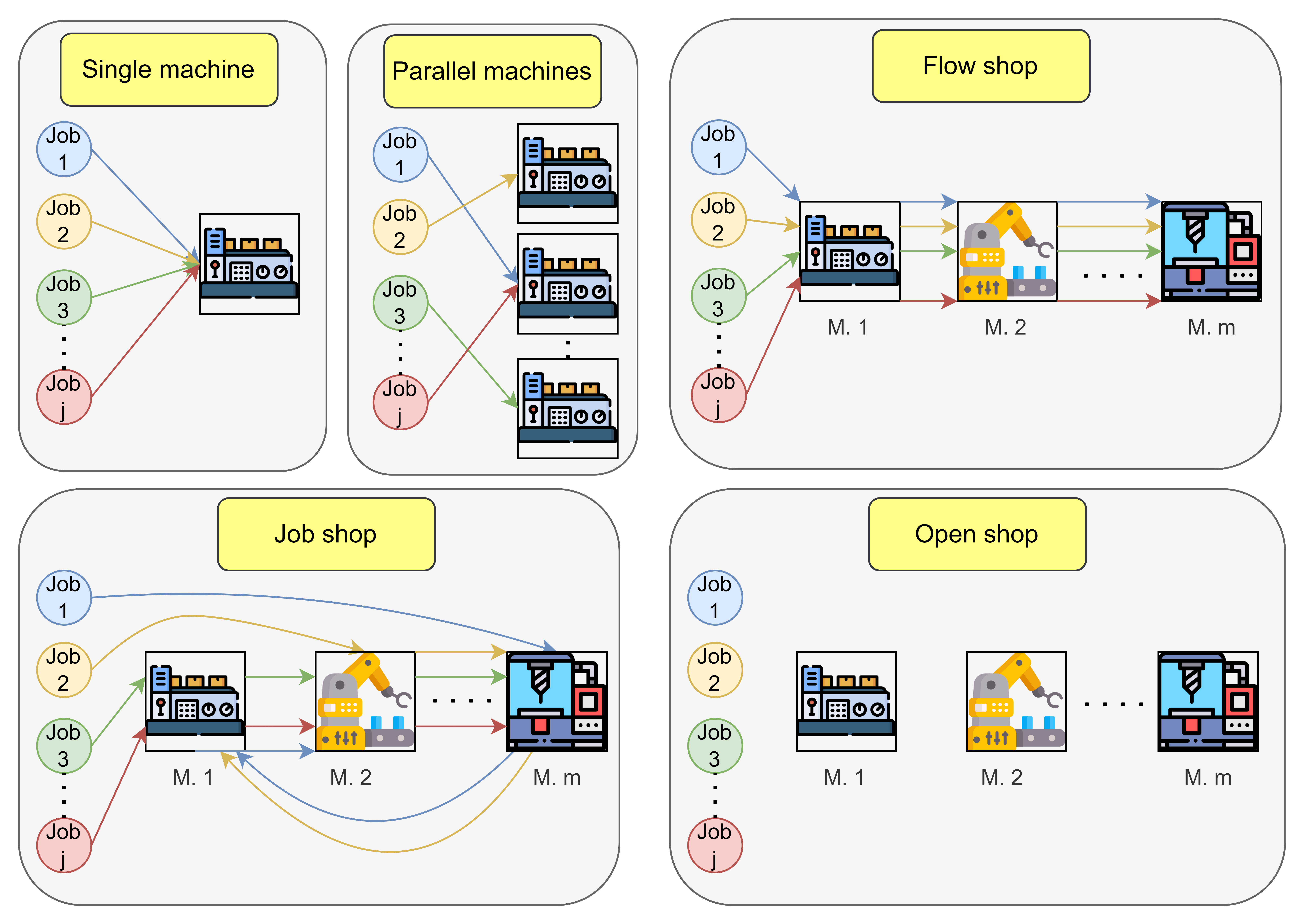}
    \caption{A classification of machine scheduling environments (J and M refer to the number of jobs and machines, respectively.}
    \label{fig:Machine}
\end{figure}

Machine environments depict how jobs are handled by machines on the shop floor, and represent the different types of scheduling problems \citep{pinedo1992scheduling}. There are five main environment types as depicted in \Cref{fig:Machine}: single machine, parallel machines, flow shop, job shop, and open shop. The \textbf{single machine} environment is the fundamental building block of all other environments. In single machine, only one machine is available and must process each job individually. The \textbf{parallel machine} environment introduces multiple machines in a parallel configuration where each job can be processed on any one of the machines. Parallel machine environments can be classified further into identical parallel machines, parallel machines with differing speeds (also called uniform parallel machines), and unrelated parallel machines. In the identical parallel machine environment, each machine operates at the same speed and is capable of processing any job. Parallel machines with different speeds is a generalization where each machine may have different operating speeds, and unrelated parallel machines is a further generalization where the operating speed depends on both machine and job. \textbf{Flow shop} environments have machines configured in series, where each job follows the same route and requires processing on each machine. Generally jobs queue through the machines in a first-in-first-out manner where no ’passing’ occurs (known as permutation flow shop). Combining parallel machines with flow shop produces an environment known as \textbf{hybrid flow shop or flexible flow shop}. In this configuration, jobs go through a series of stages where each stage can contain parallel machines. Jobs in this environment are unordered as the first-in-first-out constraint does not apply to parallel machine stages. \textbf{Job shop} is an environment of machines in which each job follows its own predetermined route. If routing requires a job to visit the same machine more than once, it is labeled as recirculation (or reentrant). By addition of parallel machines, job shop is generalized to \textbf{flexible job shop} in which each job follows its predetermined route through work centers that contain machines in parallel. As before, if routing requires a job to revisit the same work group, it is classified as recirculation. In the final environment type, \textbf{open shop}, there are no constraints on the processing order of each job, so the scheduler is able to to determine job routes and make changes. In this environment, jobs may visit and return to any machine, and some of the processing times may even be zero.

\subsection{Job characteristics} \label{sec: job_char}
Job characteristics represent constraints and processing rules of each job in the scheduling problem. These characteristics may be related to to specific requirements of the job or to the machine itself, such as its availability or configuration. Common job characteristics are summarized in \Cref{tab:job__characteristics}.

\begin{table}[ht]
    \label{tab:job__characteristics}
    \caption{An overview of various job characteristics \citep{pinedo1992scheduling}}
\centering
\begin{tabular}{p{0.3\textwidth}p{0.6\textwidth}}
 \hline
\textbf{Job Characteristics} & \textbf{Description}  \\ \hline

Release Date                                  & The release date is a restriction on when a job is able to start its processing.                                                                                                                                                                                                               \\
Preemptions                                  & Preemptions implicate that process of a job is allowed to be interrupted and rescheduled without loss of process progress.                                                                                                                                                    \\
Precedence Constraints                        & With precedence constraints, certain processes of a job may require that other process of that job have completed before they can begin.                                                                                                                                                                           \\
Sequence Dependant Setup Times                & A sequence dependant variable that represents the changeover time between two jobs ($j$ and $k$). It includes the clean-up time of job j and setup time for job k. Subscript i may be introduced (i.e., $S_{ijk}$) if times are machine dependant.                                                \\
Job Families                                  & Job families classify the different jobs j into groups that can be processed on the same machine without delay. For a machine to process another job family, requires the clean-up time and setup time for the respective   families. \\
Batch Processing                              & Machines may be capable of processing batches of up to b jobs simultaneously. Completion time for a batch is the longest completion time of jobs within that batch ($b \in [0,\inf)$).                                                                                                       \\
Breakdowns                                    & Also known as machine availability constraints, breakdowns represents that machines may become unavailable at any given time during operation.                                                                                                                                                 \\
Machine Eligibility Restrictions             & A constraint introduced when parallel machines are not all capable of processing a job j. $M_{j}$ is the set of machines capable of processing job   j.                                                                                                                                        \\
Permutation                                  & This constraints implied that sequential order in machine queues are maintained throughout the system.                                                                                                                                                                                         \\
Blocking                                     & When there is a limited buffer between machines, it is possible that machine processing may be blocked caused by an inability to release the job due to an upstream job.                                                                                                                     \\
No-Wait                                      & The no-wait constraint implies that a job may not be able to wait between successive machines. These constraints are most commonly due to conditioning that may happen during the wait time (e.g., cooling and drying).                                                                   \\ \hline
\end{tabular}
\end{table}

\subsection{Optimality criteria}

Optimality criteria can be general or specific to individual jobs and can contain one ore more performance measures. The objective function of a scheduling optimization model is defined by the given optimality criteria. The objective function is almost always a function of completion time ($C_{i}$) of jobs. The objective function might include minimization of the lateness of jobs ($L_{j}$ = $C_{j}-d_{i}$), job tardiness ($T_{j} = \max(L_{j},0)$) or penalties for jobs that are completed passed their due date. Makespan is another important criterion defined as the completion time of the last job to finish ($C_{max}$), and hence represents the overall finish time. Similarly, maximum lateness ($L_{max} = \max(L_{1}, L_{2},..., L_{n})$) represents the worst due date violation of all jobs \citep{pinedo1992scheduling}. There are many different criteria to consider, and only some of the most important have been mentioned above.

\subsection{Static and dynamic scheduling} \label{sec:staticdynamic}
In a static setting, the information of jobs to be scheduled is readily available and the schedule is determined in advance and remains fixed during execution. In a dynamic setting, the shop floor status changes in time due to of uncertainties such as dynamic job arrivals, machine breakdowns, delay in job release dates, delivery date changes, and longer than expected processing times. Majority of manufacturing systems operate in a dynamic environment \citep{mcsweeney2020efficient, Zhao2021}. In the literature, the dynamic scheduling problems were categorized into completely reactive, predictive-reactive, and robust proactive scheduling based on how they respond to dynamic events \citep{ouelhadj2009survey}. In completely reactive scheduling, waiting jobs are dispatched in real-time according to current status of shop floor and no firm schedule exists in advance. Predictive–reactive scheduling is a widely used approach in manufacturing systems. In this approach, a prior schedule (similar to static schedule) is generated at the beginning of the scheduling horizon and revised dynamically in response to real-time events. The rescheduling can be done by generating a new schedule from scratch or repairing the old schedule. The former approach only aims to only optimize the efficiency or Key Performance Indicators (KPIs), while the latter tries to reduce deviation from the old schedule and keep the plan stable. The scheduling approach that keeps a balance between efficiency and stability is referred to as robust predictive-reactive scheduling. Lastly, the robust proactive scheduling approach generates a predictive schedule while aiming to reduce the impact of future disruptions. An example to reduce the impacts of disruptions is measuring the deviation between the completion time of jobs in the realised schedule and the actual schedule to incorporate additional time intervals in the proactive schedule \citep{ouelhadj2009survey}.

\section{Methodology}\label{sec:method}
In this section, we give the definitions of MDP, which includes the states, actions, rewards, and transition functions. We also explain what the policy of an agent is and what the optimal policy is. Then, we will provide the taxonomy and basics of the most popular Deep Reinforcement Learning (DRL) techniques. In addition, we explain two widely used variants of DRL, i.e. multi-agent and hierarchical DRL, that are used in the machine scheduling literature. This section ends with explaining the advanced neural networks, including encoder-decoder architectures and Graph Neural Networks (GNN), that are used as approximations of value functions and policy functions in scheduling problems with large action and state spaces.

\subsection{Markov decision process}
\begin{figure} [H]
    \centering
    \includegraphics[width=10cm]{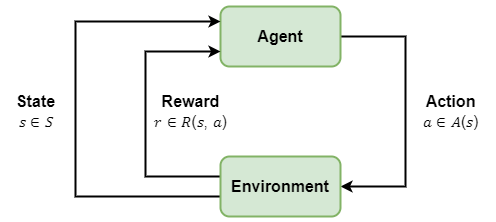}
    \caption{The Markov decision process associated with reinforcement learning \citep{sutton2018reinforcement}}
    \label{fig:mdp}
\end{figure}

To apply Reinforcement Learning (RL) methods, first the problem must be defined as a sequential decision-making process with states and actions determining an agent's ability to interact with its environment. The problem can be modeled as a Markov Decision Process (MDP) \citep{bellman1957markovian}, a common notation for modelling such problems mathematically. Figure \ref{fig:mdp} shows the structure of an MDP, representing the agent-environment interaction. The sets of states ($s \in S$) and actions ($a \in A(s)$) along with the transition dynamics ($T(s,a)$, denoted as transition probabilities $p(s',r|s,a)$) define the MDP \citep{sutton2018reinforcement}. The transition dynamics can be stochastic or deterministic, and characterize how the environment responds to actions taken by the agent. For each action the agent takes, the environment response produces a new state, $s'$, and a reward value, $r \in R(s,a)$, representing the quality of the agent's action. This reward is generally a user defined function that represents the overall goal of the agent. In an MDP problem, the agent learns a policy function, $\pi$, which maps each state to an action. The goal is to learn the optimal policy, $\pi_*$, which maps each state to the corresponding best possible action that the agent can take. The optimal policy determines the actions that maximize the expected cumulative sum of discounted rewards. The MDP can be written as a tuple: $M = <S, A, T, R, \gamma>$, and expressed mathematically with \Cref{eq:t0}.
\begin{equation}\label{eq:t0}\max_{\pi}\lim_{T\to\infty}\mathbb{E}\bigg[\sum_{t=0}^{T}\gamma^{t}r_{t}|s_{0}=s,\pi\bigg]
\end{equation}
where $\gamma \in (0,1)$ is a discount factor weighting the importance of future rewards and is introduced to ensure convergence. 

\subsection{Reinforcement learning} \label{sec:DRL}
Before discussing Deep Reinforcement Learning (DRL) solutions and state-of-the-art methods, we should first understand the tabular Reinforcement Learning (RL) methods that these more complex approaches are derived from. These preliminary techniques for solving MDPs include Dynamic Programming, Monte Carlo methods, and Temporal Difference methods \citep{sutton2018reinforcement}.

\subsubsection{Tabular reinforcement learning}
\begin{figure} [H]
    \centering
    \includegraphics[width=\linewidth]{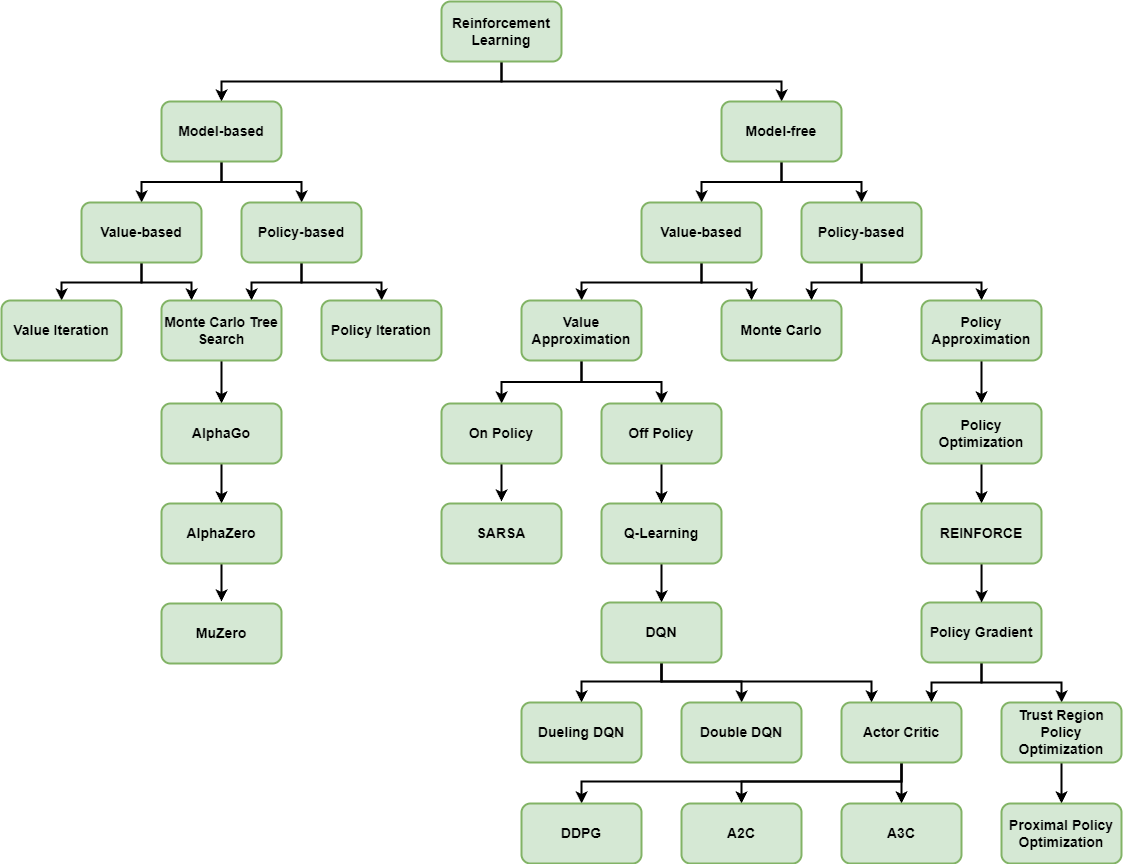}
    \caption{A classification of deep reinforcement learning methods (adopted from \citep{mazyavkina2021reinforcement})}
    \label{fig:typesofrl}
\end{figure}

RL methods can be classified into to two categories, model-based and model-free (\Cref{fig:typesofrl}). Model-based methods require knowledge of transition dynamics which are used in the algorithm's decision-making process.  These assume a closed form conditional probability of any disturbances given state-action pairs, along with defined reward and state-transition functions to solve the problem analytically \citep{bertsekas2019reinforcement}. On the other hand, model-free methods rely solely on experience and do not require environment transition knowledge. For that reason, model-free methods are more applicable for many problems, especially with real-world challenges that have unknown uncertainty. These methods generally use simulations to gain experience and estimate a solution.

To further classify RL methods, there are two general forms of approach: value based and policy based. In value based methods, the agent calculates the expected reward for the policy $\pi$, state $s$, and selected action $a$ as action values $Q^{\pi}(s,a)$ or as a state-value function $V^{\pi}(s)$, and then updates the policy to select actions that maximize the value. The optimal policy can be derived from the optimal action-value function by greedily selecting the actions corresponding the the maximal values, or from the optimal state-value function with a one-step search approach to select actions corresponding to the maximum values. Policy-based methods directly estimate a policy $\pi_{\theta}(s)$ using experience collected from the agent's previous actions. In this way, the policy is determined by maximizing the expected sum of rewards \citep{mazyavkina2021reinforcement}.

\textbf{Dymanic Programming-} Dynamic programming covers a range of model-based algorithms that exploit the known environment model to dynamically compute the optimal policy. As mentioned, these algorithms can be policy or value oriented, known as policy iteration or value iteration, based on how they iteratively update until an optimal solution is obtained. In policy iteration, a value function ($v_{\pi}$) is first computed to evaluate the current policy (\Cref{eq:t1}). This computation, termed policy evaluation, iteratively updates the value function for each state until numerical consistency is achieved.
\begin{equation}\label{eq:t1}
    v_{k+1}(s) = \mathbb{E}\big[R_{t+a}+\gamma v_{k}(S_{t+a})|S_{t}=s\big]= \sum_{a}\pi(a|s)\sum_{s',r}p(s',r|s,a)\big[r + \gamma v_{\pi}(s')\big]
\end{equation}
Next, the policy's value is compared to other actions for each state to see if a policy change is beneficial (policy improvement). This is done by considering an action $a \neq \pi(s)$ for some state $s$ and calculating the value of taking that action and then following the policy $\pi$ thereafter. The value can be calculated with \Cref{eq:t2} and then compared with the current policy's value.
\begin{equation}\label{eq:t2}
    Q_{\pi}(s,a) = \mathbb{E}\big[R_{t+1}+\gamma v_{\pi}(S_{t+1})|S_{t}=s, A_{t}=a\big]= \sum_{s',r}p(s',r|s,a)\big[r + \gamma v_{\pi}(s')\big]
\end{equation}
If the calculated value is greater than the current policy's value for state $s$, then a policy update to select $a$ whenever state $s$ occurs is justified. Now, by considering all states and all actions, this concept can be extended to select the best possible action for each state with \Cref{eq:t3}.
\begin{equation}\label{eq:t3}
    \pi'(s,a) = \arg\max_{a} Q_{\pi}(s,a) \quad \forall s \in S
\end{equation}
Repeating these processes iteratively until no greater value can be obtained is the key concept of policy iteration. In a similar fashion, value iteration combines both policy evaluation and policy improvement to reduce the computation required. Value iteration works by reducing the iterative component of policy evaluation to only a single sweep of the states and then performing the policy improvement at each step (\Cref{eq:t4}).
\begin{equation}\label{eq:t4}
    v_{k+1}(s) = \max_{a}\sum_{s',r}p(s',r|s,a)\big[r + \gamma v_{k}(s')\big]
\end{equation}

\textbf{Monte Carlo-} In contrast to Dynamic Programming, Monte Carlo methods require no underlying assumptions of the environment transition dynamics and are hence considered model-free approaches \citep{ibrahim2021applications}. These approaches use numerous simulations to calculate expected reward values and state transitions without the need of an explicit mathematical model \citep{sutton2018reinforcement}. This is an episodic process that updates value estimates and the corresponding policy at the end of each episode, meaning that experience must be separable into individual episodes.

\textbf{Temporal Difference-} Temporal Difference (TD) methods can be thought of as a combination of Dynamic Programming and Monte Carlo methods in a model-free implementation. Rather than waiting until an episode terminates as Monte Carlo does, TD methods use the experience of each time step to update an approximate solution. Similar to dynamic programming, TD methods can be either value based (value approximation) or policy based (policy approximation). There are many different TD algorithms, including Q-Learning \citep{watkins1989learning} and SARSA \citep{rummery1994line}. Q-learning is an off-policy TD method that estimates action values for each state using the transitions of state-action pairs. The algorithm starts each episode by initializing a state and choosing an action from the policy. The policy is updated using the received reward and the maximum action value for the next state. Q-Learning iteratively updates the action-value function by learning from collected experiences of the current policy (convergence to the optimal policy is proven by \citep{sutton1988learning}). SARSA is quite similar, however it follows an on-policy approach, selecting the next actions before the policy update. Both algorithms repeat this process until the episode ends for many different episodes to reach an approximate solution. Unfortunately, due to the maximization process, both of these algorithms suffer from a positive bias that overestimates the values. This is even more apparent in Q-learning \citep{tata2021investigation}, with both action selection and evaluation using the same value function. To solve this problem, \citep{hasselt2010double} introduces an improvement on the Q-Learning algorithm, called Double Q-Learning, that uses two separate value functions to reduce the maximization bias.

\textbf{Policy Gradient-} Other than policy-based TD methods, another branch of policy-based RL consists of Policy Gradient methods that make use of the policy function's gradient. The REINFORCE algorithm \citep{williams1992simple} is a precursor of Policy Gradient methods \citep{wang2020deep}. REINFORCE uses statistical methods to make adjustments along the gradient direction without computing or estimating the gradient explicitly. Conversely, Policy Gradient requires the gradient and optimizes the policy parameters, $\theta$, with a gradient descent algorithm. A standard Policy Gradient approach \citep{sutton2000policy} estimates the gradient of the policy function as shown in \Cref{eq:t5}.
\begin{equation}\label{eq:t5}
    \nabla_{\theta}J(\pi_{\theta}) = \mathbb{E}_{\pi_{\theta}}\left[\sum_{t=0}^{H}\nabla_{\theta}log\pi_{\theta}(a_{t}|s_{t})\hat{A}(s_{t},a_{t})\right]
\end{equation}
Where $H$ is the agent's horizon and $\hat{A}(s_{t},a_{t})$ is the return estimate calculated with \Cref{eq:t6}.
\begin{equation}\label{eq:t6}
    \hat{A}(s_{t},a_{t})=\sum_{t=t'}^{H}{\gamma^{t'-t}r(s_{t}',a_{t}')-b(s_{t})}
\end{equation}
Here, $b(s)$ is the baseline function which tries to mitigate any initial poor performance due to the initialization of the parameters by reducing the variance of the return estimate. 

\textbf{Actor-Critic-} Actor-Critic methods combine a policy-based approach (actor) to estimate the policy with a value-based approach (critic) to evaluate the policy for updating learnable parameters. 
The Actor-Critic approach extends REINFORCE with bootstrapping for the baseline (updating the state-value estimates from the values of following states) \citep{mazyavkina2021reinforcement}. This approach can often reduce variance further, but also introduces bias to the gradient estimates. Actor-Critic methods can be applied to continual and online learning as there is no reliance on Monte-Carlo rollouts (unrolling the trajectory to the terminal state).

\subsubsection{Deep reinforcement learning} Tabular RL frameworks may not be able to effectively solve many complex real-world systems with high-dimensional state and action spaces. This challenge led to recent research attempts in developing DRL algorithms, which adopt Deep Neural Networks (DNN) for approximating and learning policy and value functions in policy optimization. In value-based optimization, the gradient-based methods are introduced for leveraging deep neural networks, like Deep Q-Networks. In policy-based optimization, the deterministic policy gradient and stochastic policy gradient are introduced. The combination of value-based and policy-based optimization produces the popular actor-critic structure.

One of the first successful implementations of DRL is TD-Gammon \citep{tesauro1995temporal}, an algorithm developed by Gerald Tesauro in 1992. The algorithm uses a neural network trained with the temporal difference method to play the game of backgammon. Major Developments of DRL picked up much later after DeepMind revolutionized the field in 2013 with one of the first successful implementations of a DQN, capable of playing many Atari 2600 games with human-level control \citep{mnih2013playing, mnih2015human}. This work addressed instability issues of function approximation techniques and excited the community with the ability to learn from high dimensional observations, kickstarting a revolution of advancements in the field of DRL \citep{arulkumaran2017deep}. Since then, DeepMind has continued to excite the community with many other big developments, including Double-DQNs \citep{van2016deep}, AlphaGo \citep{silver2016mastering}, AlphaZero \citep{silver2018general}, and MuZero \citep{schrittwieser2020mastering}.

DQNs work similar to Q-learning, except that the tabular Q-value is replaced with a DNN as a function approximator. DQNs Commonly use an experience pool called a replay buffer to sample trajectories from when updating the network. Double-DQN extends the initial tabular implementation of Double Q-Learning with DNNs as estimators for both value functions. AlphaGo is a Monte Carlo tree search algorithm that uses a neural network trained to play the highly complex game Go. AlphaZero improves upon this method with better generalization for other tasks, and MuZero further improves AlphaZero with better performance.

Other notable DRL algorithms include variations of Policy Optimization and Actor-Critic methods. Deep Policy Gradient Methods attempt to approximate the policy's gradient with DNNs, however the algorithms' actual behaviour deviates from this motivation \citep{ilyas2018closer}. Proximal Policy Optimization (PPO) combines ideas from Policy Gradient and Trust Region Methods \citep{schulman2017proximal}, performing policy updates with constraints on the policy space \citep{mazyavkina2021reinforcement}. Deep Actor-Critic methods use separate DNNs as the actor and critic. The actor network approximates the policy and the critic network approximates the value function. Similarly, Deep Deterministic Policy Gradient (DDPG) uses the Actor-Critic structure to learn an approximate state-action function and calculate the bootstrapped return estimate. The Actor-Critic approach is extended further with Asynchronous Advantage Actor-Critic (A3C) and the synchronous version, Advantage Actor-Critic (A2C) \citep{mnih2016asynchronous}. A3C uses an asynchronous gradient decent method to optimize the DNN. A3C agents act synchronously on multiple parallel instances, eliminating the need of an experience replay buffer. This reduces the correlation of experiences and in turn leads to better stability when training. Unlike A3C, A2C lets each actor finish its experience before updating.

\subsubsection{Multi-agent deep reinforcement learning}
Multi-Agent Deep Reinforcement Learning (MADRL) allows several agents to simultaneously learn how to perform a complex task by interacting with the same environment. It offers a solution to the issue that many real-world problems cannot be effectively tackled by a single active agent that interacts with the environment \citep{stone2000multiagent}. Non-stationary and scalability are two major challenges that must be taken into consideration when developing a MARL algorithm \citep{zhuang2019scalability}. In a MADRL system, the interaction between agents, depending on the kind of reward given by the environment, can be fully cooperative, fully competitive, or a combination of both. A straightforward method for developing a MADRL system is to train all agents independently. Several studies have been conducted to investigate approaches for addressing the stability issues of a MADRL system where each agent is allowed to behave and learn independently. Among the techniques proposed in the literature are embedding other agents' policy parameters into the Q function network \citep{tesauro2003extending}, directly adding the iteration index to the replay buffer, and using importance sampling \citep{foerster2017stabilising}. To stabilize the training process of multiple agents, a centralised critic-decentralised actor that was proposed in \citep{foerster2018counterfactual} and \citep{lowe2017multi} can be used.

\subsubsection{Hierarchical deep reinforcement learning}
Hierarchical DRL (HDRL) is an approach that extends DRL methods to solve more complex problems by decomposing them into smaller problems based on a hierarchy. It is beneficial for problems with a large state-action space. HDRL algorithms outperformed standard DRL in several problems such as continuous control, long-horizon games and robot manipulation with improved exploration using subtasks \citep{pateria2021hierarchical}. There are two main approaches for learning hierarchical policies. The feudal hierarchy \citep{vezhnevets2017feudal}, uses subgoals for representing different subtasks. The framework consists of manager and worker modules, the manager sets abstract goals which are delivered to the worker, and the worker generates actions. This approach can scale up the subtask space by using many subgoals or using a low-dimensional continuous subgoal space. In policy tree approaches \citep{pateria2021hierarchical}, also known as options framework, the learning process is modeled as a semi-MDP problem. The action space of a higher-level policy consists of the different lower-level policies of subtasks. These approaches are not constrained to learning only subgoal-based subtasks \citep{sutton1999between}. The agent at the upper level chooses an action first at each stage, the final state is then examined against a set of initiation conditions. An agent at a lower level takes over the task if the starting condition is met \citep{Yan2021}. As the upper level agent can ignore implementation details, HDRL enhances exploration, and agents have higher sampling efficiency. Moreover, due to possible similarities between low-level actions, they can be transformed into various tasks within the same learning process \citep{Yan2021}.\\

\subsection{Encoder-decoder architectures} \label{sec: encoder-decoder}
The machine scheduling problem is a branch of of combinatorial optimization problems because scheduling can be formulated as an integer constrained optimization with integer or binary variables (called decision variables) \citep{Bengio2021, Liang2022}. Many combinatorial optimization problems can be converted to multi-stage decision-making problems, which means a sequence of decisions need to be made to maximize/minimize the objective function \citep{dong2022minimizing}. For instance, in machine scheduling problems, the goal is to decide the processing sequence of jobs on machines such that a criteria is optimized. This goal is similar to machine language translation where the sequence of words in one language (e.g., English) is rearranged to deliver the same meaning when is translated to another language (e.g., French) \citep{Luo2021}. This commonality between the nature of machine translation and that of combinatorial optimization problems has been a key motivation in operations research society to adopt encoder-decoder architectures, which were originally developed for language translation, in combinatorial optimization problems. These architectures consist of an encoder and a decoder to process the input data and output the optimal sequence. The encoder-decoder architectures are mainly composed of Recurrent Neural Networks (RNN); thus, we first explain the standard RNN and its extensions. Next, the encoder-decoder architectures are presented.

\begin{figure} [H]
    \centering
    \includegraphics[width=\linewidth]{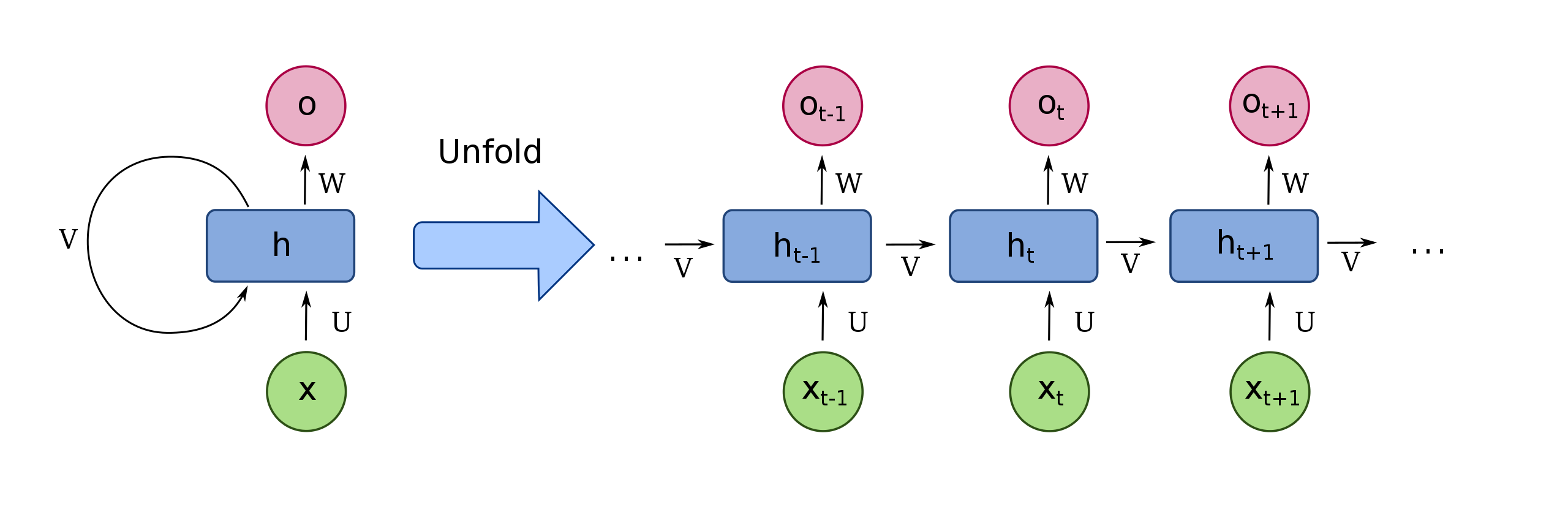}
    \caption{Architecture of standard RNN \citep{WikimediaCommons2017}}
    \label{fig:rnn}
\end{figure}

\textbf{Recurrent neural network-} The Recurrent Neural Network (RNN) is a generalization of a feedforward neural network to sequences. They are designed to process data that are naturally presented in a sequence. Signal processing, speech recognition, and machine translation are the common examples where the data is represented as a sequence \citep{Vinyals2015}. As depicted in \Cref{fig:rnn}, a standard RNN generates a sequence of outputs $(o^{1},...,o^{T})$ from the given sequence of inputs $(x^{1},...,x^{T})$ by iterating the following equations:

\begin{equation}
    h_t=\sigma\left(Ux_t+Vh_{t-1}\right)
    \label{eq:1}   
\end{equation}

\begin{equation}
    o_t=Wh_t
    \label{eq:2}   
\end{equation}

Parameter $\boldsymbol{h}$ is the hidden state variables in the RNN, while $\boldsymbol{U,V,}$ and $\boldsymbol{W}$ are the learnable parameters. $x$ and $o$ denote the input data and the output from the model, respectively. The activation function $\sigma\left(\alpha\right)$, which is used in \Cref{eq:1}, is a sigmoid function that facilitates learning the non-linear patterns in the data:

\begin{equation}
    \sigma\left(\alpha\right)=\frac{{exp}^{(\alpha)}}{{exp}^{(\alpha)}+1}
    \label{eq:3}   
\end{equation}

Since the RNN is only able to memorize short-term relationships, it has limited capacity to carry information over the network given by an input sequence with long range temporal dependencies. The Long Short-Term Memory (LSTM) and Gated Recurrent Unit (GRU) are extensions to the RNN that were developed to solve the above limitation. An LSTM utilizes three gates, namely input, output, and forget gates to determine how much information from a hidden memory cell should be exported as the output and carried over to the next hidden state \citep{Hochreiter1997}. The GRU is a simplified version of LSTM with fewer learnable parameters \citep{Chung2014}. A GRU requires less memory, has faster speed than LSTM and exhibits better performance in certain less frequent data sets with short sequence. Readers can find the mathematical formulation of LSTM and GRU networks in \citep{Hochreiter1997} and \citep{Chung2014}, respectively.\\

\begin{figure} [H]
    \centering
    \includegraphics[width=15cm]{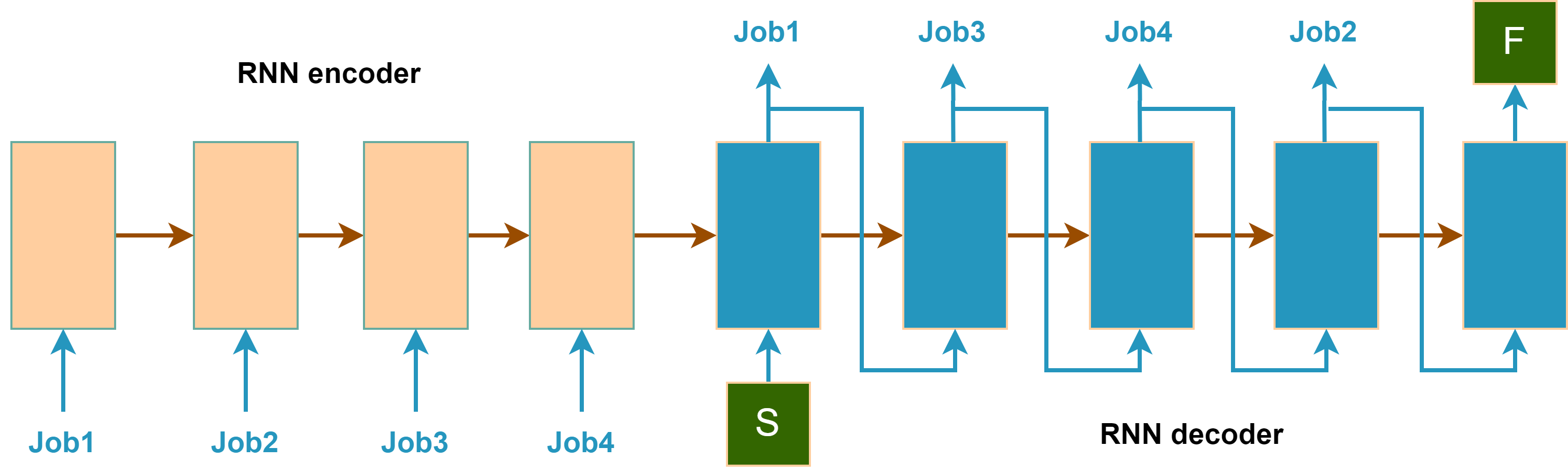}
    \caption{Sequence-to-Sequence architecture (adopted from \citep{dong2022minimizing, Sutskever2014})}
    \label{fig:seq2seq}
\end{figure}

\textbf{Sequence-to-sequence model- } Sequence-to-sequence learning (Seq2Seq) is the first developed encoder-decoder architecture. It was proposed to address the limitation of a single RNN in mapping a fixed-dimensional input to a fixed-dimensional output of potentially different size when the dimensions of input and output are not known in advance \citep{Sutskever2014}. The Seq2Seq model utilizes an RNN to encode the sequential input to a fixed-length feature vector representing the abstract information of the input sequence. Then, another RNN decodes the feature vector to the target sequence. \Cref{fig:seq2seq} demonstrates the overall architecture of a Seq2Seq model for machine scheduling. An RNN (pink) encodes the input sequence of nodes to a feature vector that is used by the RNN of the decoder (blue) to generate the output sequence.

The goal of Seq2Seq model is to calculate the conditional probability $p\left(C^P|P;\theta\right)$, given a training pair $\left(P;C^P\right)$. As depicted in \Cref{eq:4}, the conditional probability can be factorized according to the chain rule. The factors in \Cref{eq:4} can be estimated using a trainable model such as an RNN with parameters $\theta$.

\begin{equation}
    p\left(C^P|P;\theta\right)=\prod_{i=1}^{m\left(P\right)}{p_\theta\left(C_i|C_1,\ldots,C_{i-1},P;\theta\right)}
    \label{eq:4}   
\end{equation}

Here $P=\left\{P_1,{\ldots,P}_n\right\}$ denotes an input sequence of $n$ vectors, while $C=\left\{C_1,\ldots,\ C_{m\left(P\right)}\right\}$ is an output sequence of $m\left(P\right)$ indices corresponding to positions in $P$, each between 1 and $n$. The learnable parameters of the RNN are computed to maximize the conditional probabilities for the training data set through the equation below.

\begin{equation}
   \theta^\ast={arg}_\theta max\sum_{P,C^P}\log{p\left(C^P|P;\theta\right)}
    \label{eq:5}   
\end{equation}

\textbf{Attention mechanism- }\label{sec: attention_mech} The Seq2Seq architecture encodes all the information available in the input sequence $P$ to a fixed-length feature vector. The fixed-sized vector may not be able to provide all the necessary information to the decoder when generating the output sequence $C^P$. This limitation causes difficulty when dealing with long inputs, especially those that are longer than the sequences observed during the training. To solve this issue, \citep{Bahdanau2015} augmented the RNNs of encoder and decoder with an additional neural network. The neural network adoptively decides to pay attention to a subset of the hidden states of the RNN encoder that are most relevant to generating a correct output during decoding. Assuming the hidden states of encoder and decoder RNNs are denoted as $\left(e_1,\ \ldots,\ e_n\right)$ and $\left(d_1,\ \ldots,\ d_{m\left(P\right)}\right)$, respectively, the attention distribution $a^i$ at each decoding time step $i$ is computed through \Cref{eq:6} and \Cref{eq:7}. Note that $v, W_1,$ and $W_2$ are learnable parameters that help to augment the RNNs of the encoder and decoder. 

\begin{equation}
   u_j^i=\ v^T\ tanh\left(W_1e_j+W_2d_i\right)
   \qquad
   j\in\left(1,\ \ldots,\ n\right)
    \label{eq:6}   
\end{equation}

\begin{equation}
   a_j^i=softmax\ \left(u_j^i\right)=\ \frac{exp\left(u_j^i\right)}{\sum_{k=1}^{n}exp\left(u_k^i\right)}
   \qquad
   j\in\left(1,\ \ldots,\ n\right)
    \label{eq:7}   
\end{equation}

Using softmax function in \Cref{eq:7}, the terms in vector $u^i$ are normalized to obtain the attention distribution (also called attention function). The attention distribution guides the decoder at each time step $i$ to decide which part of the input sequence to concentrate on when it generates the next element of the output sequence \citep{Esmaeilzadeh2019}. Next, the attention distribution is used to calculate the weighted sum of the encoder hidden states at time step $i$, $d_i^\prime$, known as the context vector (see \Cref{eq:8}). The context vector represents what has been read from the encoder hidden states in time step $i$. Lastly, $d_i^\prime$ and $d_i$ are concatenated to make predictions at current time step and to be fed as the hidden states of the next time step of the decoder recurrent model \citep{Bahdanau2015}.

\begin{equation}
    d_i^\prime=\ \sum_{j=1}^{n}{a_j^ie_j}
    \label{eq:8}   
\end{equation}

\begin{figure} [H]
    \centering
    \includegraphics[width=15cm]{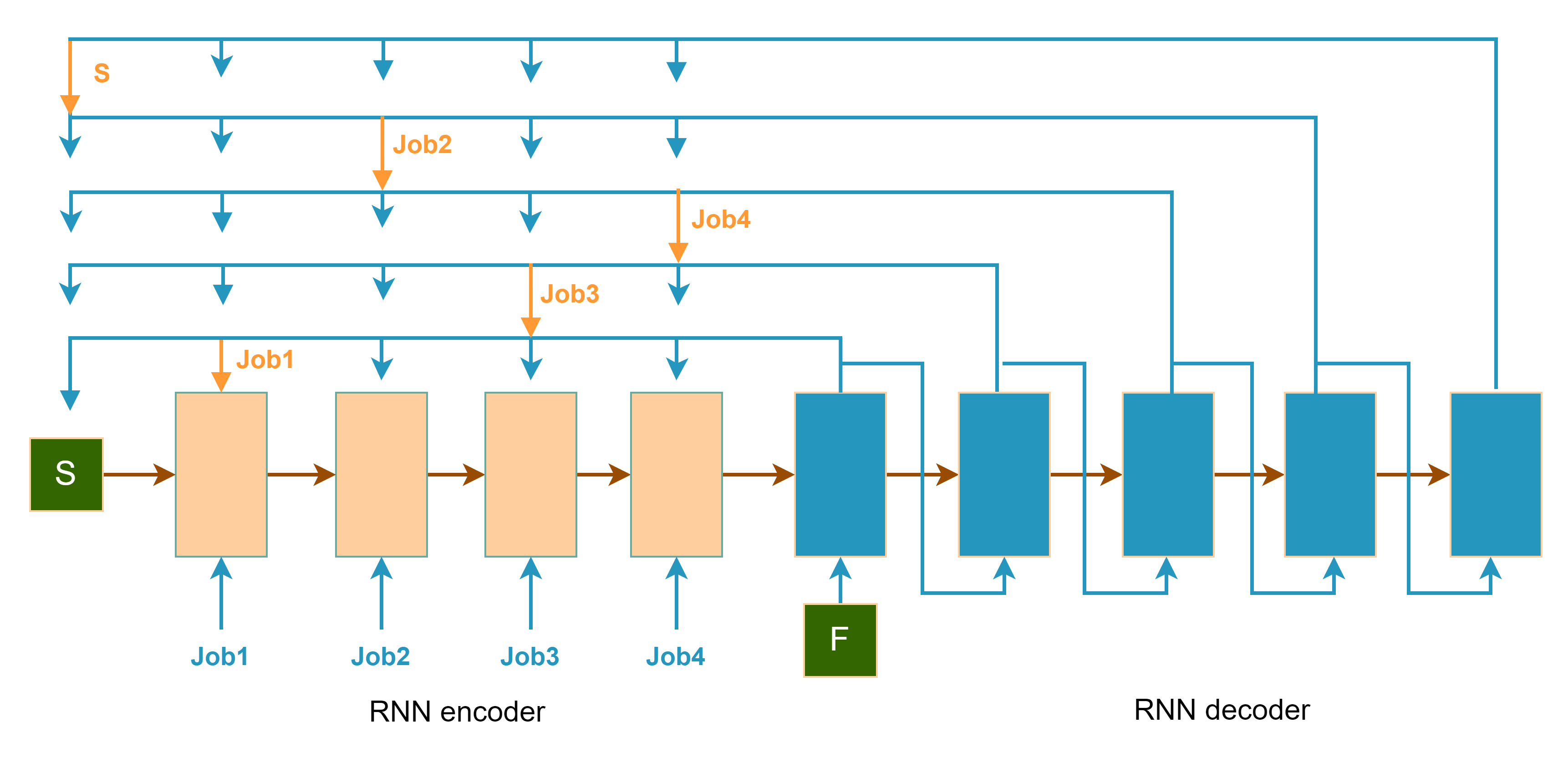}
    \caption{Pointer network architecture (adopted from \citep{dong2022minimizing, Vinyals2015})}
    \label{fig:ptr}
\end{figure}

\textbf{Pointer network- }The Seq2Seq model (with or without attention mechanism) only performs well on sequential problems with fixed-sized inputs and outputs. In combinatorial optimization problems, the size of the input varies in each instance, yielding variable sized outputs. When the problem size varies, separate Seq2Seq models should be trained for each problem size which is not computationally efficient. To overcome this limitation, \citet{Vinyals2015} developed a size-agnostic model, called Pointer Network (PN) by a modification (reduction) in the architecture of "Seq2Seq with attention mechanism". They used the attention distribution as pointers to the element of inputs. This technique effectively allows the model to point to a specific position in the input sequence at each decoding step instead of predicting an index value from a fixed dictionary size. The input element with the highest probability is selected to be decoded in each decoding step following \Cref{eq:9}. Since the softmax function in \Cref{eq:9} (that normalizes the vector $u^i$ of length $n$) is used to directly calculate the conditional probabilities, it enables the model to handle input sequences with different lengths. The PN also eliminates the calculation of the context vector in \Cref{eq:8}; thus, it does not require blending the encoder and decoder hidden states to provide extra information to the RNN of decoder. \Cref{fig:ptr} demonstrates the overall architecture of a PN for solving scheduling problems. After processing the input sequence with the encoder (pink). The decoder (blue) points to an input element to be outputted at each decoding step (orange arrows). In the PN, a glimpse function can be introduced prior to the calculation of the attention function to aggregate the contributions of different parts of the input sequence \citep{vinyals2015order}. Similar to the attention function, the glimpse function is calculated through \Cref{eq:6,eq:7,eq:8}. Then, the output of the glimpse function together with the encoder hidden states are used in \Cref{eq:6,eq:7} to calculate the attention function.

\begin{equation}
    P\left(C_i|C_1,\ldots,C_{i-1},P\right)=\ softmax\ \left(u^i\right)
    \label{eq:9}   
\end{equation}

\begin{figure}[H]
    \centering
    \includegraphics[width=10cm]{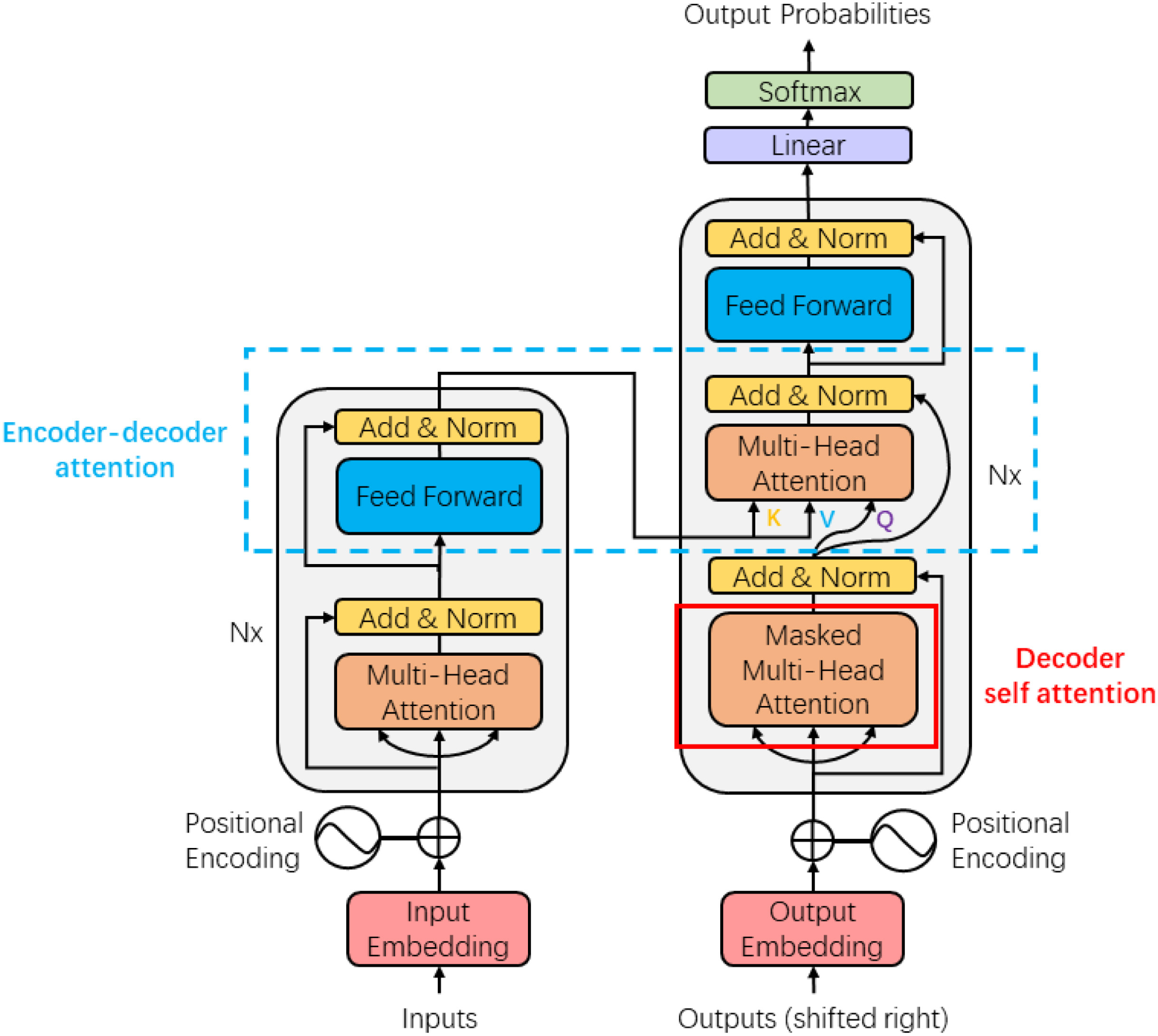}
    \caption{Transformer network architecture  (adopted from (\citep{vaswani2017attention})}
    \label{fig:transformer}
\end{figure}

\textbf{Transformer network-} \label{sec:Transnet}The sequential nature of the recurrent networks inhibits parallelization, resulting in the increase of the computation time. This limitation motivated \citep{vaswani2017attention} to develop the transformer network that has an encoder-decoder structure similar to those of Seq2Seq model and pointer network; however, it does not rely on the recurrent networks. Instead, it utilizes a stack of identical multi-head self-attention layers and position-wise fully connected feedforward neural networks in both the encoder and decoder structure. \Cref{fig:transformer} demonstrates the architecture of a transformer network. The embeddings of the input sequence along with their positional encoding are fed to the multi-head attention layers. At each attention layer (also called ‘head’) of the encoder, three different vectors, including key (K), query (Q), and value (V) are computed with three different matrices of weights to represent the input. These learnable parameters are used to calculate the attention that should be paid to each element in the sequence following \Cref{eq:10}.

\begin{equation}
    Attention\left(Q,K,V\right)=softmax\left(\frac{QK^T}{\sqrt{d_K}}\right)V
    \label{eq:10}   
\end{equation}

According to \Cref{eq:10}, the dot product between the ‘query’ of each element (e.g., input $i$) in the input sequence and the ‘key’ of other input elements are first calculated and scaled by $(\frac{1}{\sqrt{d_k}})$, where $d_k$ denotes the dimension of ‘key’ matrix. Then, the softmax of the key-query products are calculated and multiplied by the ‘value’ of the input elements to calculate the amount of attention that should be paid to the input element $i$ by other input elements. The goal of this operation is to pull up the most useful information required for computing the best representation of each input element. As depicted in \Cref{fig:transformer}, the attention vector of all multi-head self-attention layers are concatenated and fed to a fully connected neural network for extracting the most important information out of the attention values. The aforementioned steps are repeated $n$ times. Through this repetition, self-attention occurs in which the key, query, and values from step $n-1$ of attention are carried over to the next step of the attention ($n$).\\
A similar operation takes place in the decoder with a difference in that the key and value calculated by the encoder are still used along with the query computed by the decoder to obtain the attentions. Based on the softmax of the decoder attentions, the probability of each element to appear as output in current encoding step is calculated. There are three advantages associated with the transformer network over the RNN-based encoder-decoder models, including: less complexity over each layer, parallelization capability, and the ability to learn long-range dependencies in the input sequence.

\subsection{Graph neural networks} \label{sec: GNNs}
Graphs are data structures that are widely employed to capture the structural information of various problems such as social networks, recommender systems, scientific citations, molecular graph structures, as well as combinatorial optimization \citep{wu2021graph}. Nodes are representatives of the units in the data structure, while edges show the interaction among the units. The key assumption behind graph structured data is that there exist meaningful relations among the graph components, that if discovered can provide valuable insight into the data and can be utilized for the succeeding machine learning tasks such as node classification and edge prediction \citep{vesselinova2020learning}. Despite text, image, and audio, graph structured data do not have a clear grid structure, causing difficulty for convetional feedforward, convolutional, or recurrent neural networks to properly accommodate the sparse structure of graphs in their frameworks. As a result, they have a limited ability in learning the structural information behind the graphs \citep{zhang2020deep}. Graph Neural Networks (GNN) have been developed to overcome this limitation. They are able to learn often complex relations among nodes and the rules that govern these relations by processing the input graph \citep{vesselinova2020learning}. Having the node and edge attributes along with the graph structure as inputs, GNNs can be utilized for three different analytic tasks: (1) node level, (2) edge level, and (3) graph level task. At node level, the embeddings calculated for each node by GNNs are used for node classification and node regression. The embedding of nodes that form the head and tail of an edge are used for edge level tasks such as link prediction and edge classification. For graph level tasks such as graph classification, the embeddings of nodes are fed as inputs to a "readout function" to generate a feature vector representing the whole graph \citep{wu2021graph}. The readout function is explained in \Cref{sec: readout}. In the remainder of this section, we first briefly review the graph definition and explain popular graphs applied to machine scheduling tasks. Next, we present the GNNs that have been adopted to solve machine scheduling problems.\\

\textbf{Graph definitions-} A graph can be mathematically expressed as $G\ =\ <V,\ E>$ where $v\in V$ denotes a node (vertex) of the graph, and $e_{ij}=\left(v_i,v_j\right)\in E$ denotes an edge (arc) connecting the nodes $v_i$ and $v_j$ of the graph. In addition, each graph has an adjacency matrix, representing the connection of each node with other nodes. Assuming that the adjacency matrix of graph G to be $A=\left[A_{ij}\right]_{n\times n}$, where $n$ is equal to $\left|V\right|$, if $e_{ij}\in E$, then $A_{ij}=w_{ij}>0$, and if $e_{ij}\notin E$, then $A_{ij}=0$. The number of edges connected to a node determines the node degree, that is formally defined as $deg\left(v_i\right)=\left|N_{\left(v_i\right)}\right|$, where $N_{\left(v_i\right)}$ denotes the set of all neighbors of node $v_i$ \citep{wu2021graph}. \textbf{Complete graph}: a graph in which each node is fully connected to other nodes, is called complete graph. In a complete graph, $deg(v)=\left | n \right | \ \forall \ v\in V$. \textbf{Directed/undirected graph}: a graph is undirected only if its adjacency matrix is symmetrical, i.e., $A_{ij}=A_{ji}$ for any $1\le i,j\le n$, otherwise, the graph is directed. \textbf{Directed Acyclic Graph (DAG)}: a directed graph with no directed cycles is considered acyclic. \textbf{Attributed graph}: a graph may have either node attributes, edge attributes, or both. The node attributes are denoted as the matrix of $X\in R^{n\times d}$, where $x_v\in R^d$ represents the feature vector of node $v$. Similarly, the edge attributes are denoted as matrix of $X^e\in R^{m\times c}$, where $x_{\left(v,u\right)}\in R^c$ represents the feature vector of edge $\left(v,u\right)$. \textbf{Disjunctive graph}: it is a way of representing scheduling problems as a graph. In a disjunctive graph, the operations of jobs and timing constraints among operations are represented as nodes and directed edges of the graph, respectively. There are two type of edges i.e., conjunctive and disjunctive. The directed conjunctive edge between two nodes shows the precedence-succeeding constraint between those two nodes depending on the edge direction. The directed disjunctive edge between two nodes shows machine-sharing constraints where the edge direction determines the sequence of processing those operations on the machine.\\

\textbf{Recurrent GNN-} The first GNN architecture was developed by \citep{scarselli2008graph}, which is referred to as recurrent GNN (or simply GNN) as it is inspired by how recurrent neural networks function. The basic idea behind a GNN is to let nodes update their individual hidden state by recursive exchange of information with their neighbor nodes until their hidden states reach a stable equilibrium. The hidden state of each node is a low-dimensional vector that can be used to represent the embedding of each node once after convergence happens. \Cref{eq:11} shows the update function, where $h_v^{\left(t-1\right)}$ and $h_u^{\left(t-1\right)}$ denote the hidden states of the node $(v)$ and its neighbors $(u)$ at time step $t$, respectively. The hidden states of nodes at the first step ($h_v^{\left(0\right)}$) are initialized with random vectors. The terms $x_v,x_u$ and $x_{\left(v,u\right)}^e$ denote the attributes of the node, its neighbors, and the connecting edges, respectively. Lastly, $f()$ is a differentiable function such as a feedforward neural network that helps the hidden states of nodes to converge. After convergence, the last hidden state of each node represents its embedding \citep{vesselinova2020learning,hamilton2017representation}.

\begin{equation}
h_v^{\left(t\right)}=\sum_{u\in N\left(v\right)} f\left(h_v^{\left(t-1\right)},h_u^{\left(t-1\right)},x_v,x_{\left(v,u\right)}^e,x_u\right)
    \label{eq:11}   
\end{equation}

\textbf{Graph Convolutional Network (GCN)-} GCN is an adjusted format of convolutional neural networks for learning graph data \citep{welling2016semi}. Similar to a simple GNN, a GCN computes the embedding of each node by aggregating the attributes of the node with those of its neighbors. Despite GNN using the same function such as a feedforward neural network to update the hidden state of nodes, GCN uses a stack of convolutional layers each with its own weights. By using the element-wise mean pooling in GCN, the node embeddings are obtained through \Cref{eq:12}, where $W^k$ is the learnable weight matrix of layer $k$. The hidden state of every node is initialized with its feature vector (node attributes).

\begin{equation}
h_v^{\left(k\right)}=ReLU\left(W^k.MEAN\left\{h_u^{\left(k-1\right)},\ \forall u\in N\left(v\right)\cup\left\{v\right\}\right\}\right)
    \label{eq:12}   
\end{equation}

ReLU stands for Rectified Linear Unit. It is commonly used in neural networks as an activation function and enables neurons to capture non-linearity in the input data. It is defined as $ReLU(x)= max(0,x)$.\\

\textbf{Message Passing Neural Network (MPNN)-} MPNN is another type of popular convolution-based GNNs, proposed by \citep{gilmer2017neural}. In MPNN, each node communicates with its immediate neighbors by sending messages to them and receiving their messages to update its own hidden state. Message function $(g)$ and update function $(f)$ contain the learnable parameters of the MPNN and are calculated by \Cref{eq:13} and \Cref{eq:14}, respectively.

\begin{equation}
m_v^k=\sum_{u\in N\left(v\right)}{g^k\left(h_v^{k-1},h_u^{k-1},x_{\left(v,u\right)}^e\right)}
    \label{eq:13}   
\end{equation}

\begin{equation}
h_v^k=f^k\left(h_v^{k-1},m_v^k\right)
    \label{eq:14}   
\end{equation}

\textbf{Graph Isomorphism Network (GIN)-} GIN, although simple in structure, is proven to be the most powerful network among the class of GNNs \citep{xu2018powerful}. It updates the hidden states of nodes following \Cref{eq:15}, where $\varepsilon$ can be fixed to a scalar or considered as a learnable parameter. The term ${MLP}^{(k)}$ in \Cref{eq:15} refers to k$^{th}$ layer of a multi-layer perceptron.

\begin{equation}
h_v^{(k)}={MLP}^{(k)}\left(\left(1+\varepsilon^{\left(k\right)}\right).h_v^{\left(k-1\right)}+\sum_{u\in N\left(v\right)} h_u^{\left(k-1\right)}\right)
    \label{eq:15}
\end{equation}

\textbf{Graph attention network (GAT)-} GAT was developed by \citep{velivckovic2017graph} to generalize the concept of the attention mechanism reviewed in \Cref{sec: attention_mech} to graph data. In the initial step, a parameterized weight matrix $W$ is applied to every node in order to linearly transform their input features into higher-level features. This linear transformation raises the expressive power of the model. Next, a single-layer feedforward neural network, denoted by $a$, is used as the attention function to compute attention coefficients based on \Cref{eq:16}. The attention coefficient indicates the importance of the features of neighbor node $u$ to node $v$. In order to make the scale of the coefficients all the same, they are normalized across all immediate neighbors of node $v$ by \Cref{eq:17}.

\begin{equation}
e_{vu}=a\left(Wh_v,Wh_u\right)\ \ \ \ \ \ \ \ \ \ \ \forall\ u\in N\left(v\right)
    \label{eq:16}   
\end{equation}

\begin{equation}
a_{vu}={softmax}_u\left(e_{vu}\right)= \frac{exp\left(e_{vu}\right)}{\sum_{u\in N\left(v\right)} e x p\left(e_{vu}\right)}
    \label{eq:17}  
\end{equation}

Following \Cref{eq:18}, the normalized attention coefficients are multiplied by the features corresponding to them and their linear combination is used to serve as the final output features (embeddings) for every node. The term $\sigma$ denotes the sigmoid function that adds non-linearity to \Cref{eq:18}. See \Cref{eq:3} for the mathematical formulation of sigmoid function. 

\begin{equation}
h_v^\prime=\sigma\left(\sum_{u\epsilon N\left(v\right)}{a_{vu}Wh_u}\right)
    \label{eq:18}   
\end{equation}

The concept of multi-head attentions that is originally used in transformer networks can be employed to enhance the stability of a GAT. Through multi-head attentions, GATs can perform multiple independent attention mechanisms in parallel. When a GAT is applied to a complete graph (i.e., fully connected graph), it performs in its most general form where it drops all structural infromation and lets each node to pay attention to every other node. Under such circumstance, GATs work very similar to transformer networks that we reviewed in \Cref{sec:Transnet}.\\

\textbf{Readout function-} \label{sec: readout}In all GNN classes, the node hidden states $h_v^{\left(k\right)}$ at final iteration $k$, are used as the node embeddings for node level tasks such as node classification. For graph level tasks (e.g., graph classification), the node embeddings require being aggregated by a readout function to output the graph embedding, $h_G$. through \Cref{eq:19}. The readout function can be a sum over the node embeddings or a more sophisticated pooling function such as the mean or maximum \citep{xu2018powerful}.
\begin{equation}
h_G=READOUT\left(\left\{h_v^{\left(k\right)}|v\in G\right\}\right)
    \label{eq:19}   
\end{equation}

\section{DRL approaches for machine scheduling} \label{sec: approaches}

\subsection{Conventional DRL} \label{sec:conventinalDRL}

In this approach, the scheduling problems is formulated as a Markov Decision Process (MDP) \citep{Park2021}. The state space is defined by the properties of the jobs that are schedulable at each decision point and properties of machines that are available for processing the jobs \citep{waubert2022reliability}. At each decision point, the agent's action is dispatching the schedulable jobs to the available machines based on the current state of the shop floor \citep{zhou2022reinforcement}. A job is selected for dispatching based on multiple predefined priority-based dispatching rules (i.e., heuristics). When the agent takes a good action at a given state, it receives a reward. Through an iterative interaction with the environment, the agent gradually learns an optimal policy for dispatching right jobs to right machines at every production state \citep{Park2021}. DRL agent interacts with a simulated environment of the machine environment to learn the optimal policy for dispatching jobs to machines. Despite traditional RL methods that use look-up tables to store the state or action values, DRL models under this category leverage the approximation power of conventional neural networks including feedforward, convolutional, and recurrent neural networks for estimating the value function or policy function of the agent. Using these neural networks enables the agent to learn which action to take even in previously unseen states and helps the agent to perform well in problems with large state and action space \citep{waubert2022reliability}.

\subsection{Advanced DRL} \label{sec:advancedDRL}
Machine scheduling can be represented as a sequential decision making problem or a graph. Since advanced neural networks, including encoder-decoder architectures (\Cref{sec: encoder-decoder}) and Graph Neural Networks (GNNs) (\Cref{sec: GNNs}) are specifically designed for sequential data and graph-structured data, they can be adopted for optimizing machine scheduling problems. The encoder-decoders and GNNs can be trained either in a supervised or an unsupervised basis. Supervised learning methods should be provided with a high-quality labeled training data set to perform well, though providing such data is not an easy task in combinatorial optimization problems (herein machine scheduling). This is because of NP-hardness of the problems in this domain that makes finding the global optimal solution for labeling each training instance computationally expensive and in case of large-scale problems even impossible. To address this limitation of supervised learning methods, policy gradient-based RL methods can be used to train the learnable parameters of encoder-decoders and GNNs in an unsupervised manner \citep{vesselinova2020learning, bello2016neural}. The encoder-decoders or GNNs shape the computational component of the RL model. They approximate the value function and policy function of the agent. Since encoder-decoders or GNNs have more advanced architectures as compared to the conventional neural networks, we call the DRL models that use them as their computational component as advanced DRL.\\

In advanced DRL, similar to conventional DRL, the state space is represented by the machine and job status. First, a scheduling problem instance is modeled as a sequence or as a graph and then processed by an encoder-decoder or a GNN. The employed encoder-decoder or GNN summarises the sequential or structural information of the target problem to a low-dimensional feature vector that well represents the problem instance. This vector is used by the policy function to determine which job to be dispatched next. Unlike conventional DRL that the agent's action space consists of a certain number of priority-based dispatching rules (heuristics) to decide which job to be processed next, the agent in the advanced DRL directly selects a job (as its action) to be processed next. Consequently, advanced DRL inherently has larger action spaces. Policy gradient-based and actor-critic methods are often employed to train the agent in advanced DRL because these methods perform better in MDP problems with large action space as compared to value-based methods such as Deep Q Network (DQN) \citep{sutton2018reinforcement}. Based on the reward that is received by the RL agent, it adjusts the learnable parameters of the encoder-decoder or GNN such that they can generate optimal schedules with minimum cost (e.g. minimum makespan) \citep{Nazari2018, bello2016neural}. Encoder-decoders and GNNs enable the trained policy to handle sequences with different lengths or graphs with different sizes \citep{Park2021}. As a result, the scheduling policy can be employed to generate a schedule for problem instances with different sizes. The way that advanced DRL deals with the scheduling problems can be divided into Learn-to-Construct (L2C) and Learn-to-Improve (L2I). L2C methods build the solutions incrementally using the learned policy. At each step, they choose which element to be added to a partial solution. L2I methods start from an arbitrary solution and learn an optimal policy to improve it iteratively. The L2I methods tackle the issue that is commonly encountered with the L2C methods, namely, the need to use some extra procedures to find a good solution such as sampling or beam search \citep{mazyavkina2021reinforcement}.

\subsection{Metaheuristic-based DRL} \label{sec:metaheuriticDRL}
The scheduling problem can be considered as an NP-hard problem because the solution space becomes exponentially bigger when multiple machines and products are brought into play \citep{brucker1998scheduling}. This can rise the necessity for an efficient search in the solution space. In fact, to only find the optimum combination for production planning, $(n!)m$ would be the order of the possible combinations, where $n$ and $m$ respectively denote the number of jobs and the number of machines \citep{blum2003metaheuristics}. In the past few decades, many linear and non-linear optimization solutions are proposed in the literature to cope with the problem but they usually require considerable computational power and are ineffective when the environment is dynamic and stochastic. Later on, metaheuristic algorithms introduced a better efficacy in alleviating the computational burden and finding a near optimum solution in a timely-manner. Thus, they received an increasing interest among the scholars to deal with machine scheduling \citep{para2022energy}. However, many of these algorithms require exhaustive fine-tuning, more specifically for the number of population or other mutation-related parameters. In such a case, some recent papers have proposed frameworks in which an RL agent tries to tune these parameters in a systematic manner. In other words, after learning from multiple scenarios and samples that the metaheuristic model has solved with different parameters, the agent is trained to figure which parameter values it picks for the metaheuristic model based on the given scenario, to push it faster towards the global optima.\\

The metaheuristic model can also be employed as an optimizer to generate better solutions at each episode during the training phase of an RL agent and enhance its performance. In these frameworks a simulator is developed to provide a sandbox for the agent to train on. At each time step or episode the state, actions and the corresponding reward are calculated by the simulator. Therefore, a variety of scenarios are provided for the agent to learn from in an offline manner. The experiments that the agent faces at each time step are recorded in the memory. Using an experience replay technique, the agent can learn from these experiences over time and achieve the ability of finding good scheduling solutions in real-time. Moreover, the simulator can transform the near-optimal schedule solution generated by the metaheuristic model at each episode to the trajectory. The obtained optimal trajectory solution can be restored into the memory to enhance agent training efficiency in each episode \citep{chien2021agent}.

\section{Article collection methodology}

We carried out our literature search through Web of Science to consider publications in English language from peer-reviewed journals, proceedings, conference papers, and books. Additional literature search was conducted using Google Scholar as a complement to the articles found using Web of Science. The keyword search in Web of Science and Google Scholar was complemented by a backward and forward search in the articles found within these two websites. To ensure that we would comprehensively review the application of different DRL approaches in scheduling of different machine environments, we defined the search query and keywords in an iterative process. We first read the abstract of articles found based on an initial search query and iteratively added the missing keywords related to the scope of our literature review to the search query until no new paper was found. The final search query consisting of two-level keyword assembly is given in \Cref{tab:query}. The first level is concerned with the domain application of studies, while the second level is concerned with the solution method of studies. Both levels look up the keywords in the topic of papers (i.e., Title, Keywords, or Abstract). Domain application was assembled by considering “schedul*” OR “dispatch*” AND different machine environments reviewed in \Cref{sec: machine_env}. We considered "dispatch*" beside scheduling since within the iterative process, we noticed that dispatching and scheduling are interchangeably used in the context of machine scheduling problems. Moreover, some studies conducted scheduling on production or manufacturing systems which can be formulated as one of the machine environments discussed in \Cref{sec: machine_env}, but these studies did not point out the type of machine environment in their topic. To avoid missing this group of papers, we incorporated manufacturing, production, and factory as general keywords along with different machine environments in the first line of query.\\

The intent of second line of query is to find all the papers that applied RL as their solution approach to solve machine scheduling problems. We did not limit the keywords only to DRL algorithms and considered other RL methods including Q-learning, SARSA, TD, and Monte Carlo learning. The reason behind this inclusion is that some authors applied one of the above methods (e.g., Q-learning) to their problem and used an artificial neural network to approximate the state or action values, while they did not label their work as “deep RL”. These studies did not use the term "deep reinforcement learning" because they were conducted before when the DRL word or the terms related to DRL-based algorithms were coined. The final search resulted in 480 papers. We first screened out the papers that were out of scope of machine scheduling. These papers were mostly related to task scheduling of robots, scheduling in communication networks, scheduling of energy systems, maintenance scheduling, logistics, and supply chain management. Thereafter, a total of 72 papers remained. Each of them falls under one of the categories of conventional DRL, advanced DRL, or DRL along metaheuristic that were presented in \Cref{sec: approaches}. In the following section, the DRL-based papers are reviewed. Further analysis on the annual publication trend of different DRL approaches is provided in \Cref{sec:trend}.

\begin{table}[ht]
\centering
\caption{The proposed two-level keyword assembly structure}
\label{tab:query}
\begin{tabular}{p{1.5cm}p{8cm}p{3cm}}
\toprule
Context & Query & Searching field \\
\midrule
Application domain & ("schedul*" or "dispatch*") and ("single machine" or "parallel machine" or "flow shop" or "job shop" or "open shop" or "multi machine" or "manufactur*" or "production" or "factory") & Title \& Keywords \& Abstract \\ \\

Methodology & "reinforcement learning" or "Q-learning"  or "SARSA" or "temporal difference" or "TD" or "Monte Carlo learning" or "neuro dynamic" or "DQN" or "DDQN" or "REINFORCE" or "PPO" or "DDPG" or "policy gradient" or "actor critic" or "A2C" or "A3C" & Title \& Keywords \& Abstract\\

\bottomrule
\end{tabular}
\end{table}

\section{Application} \label{sec:applications}
In this section, we review the application of Deep Reinforcement Learning (DRL) in machine scheduling problems. The reviewed articles can be classified by two dimensions: methodology, and machine environment that we explained in \Cref{sec: approaches} and \Cref{sec: machine_env}, respectively. Hence, the articles are first categorized by their methodology and then grouped based on their machine environment. Important features of the reviewed articles under each methodology and each machine environment are provided in separate tables. The important features consist of their applied DRL algorithm, optimality criteria, the number of agents, problem size, and their performance summary.

\subsection{Conventional DRL}

In this subsection, papers that adopted conventional DRL methods to solve machine scheduling problems are reviewed. As discussed in \Cref{sec:conventinalDRL}, conventional DRL methods refer to the DRL models that use either feedforward, convolutional, or recurrent neural networks as their computational component. The main goal of this category of studies is to find the optimal policy to dispatch jobs using a set of priority dispatching rules (heuristics). These studies were grouped and reviewed according to their machine environment.\\

\textbf{Single machine-} In \citep{Xie2019}, the authors determined the best features to represent states and actions for production scheduling of jobs on a single machine, by testing a large number of experimental cases and developing several reward functions. They concluded that adding unnecessary inputs as state features only expands the state space and deteriorates model performance. As a result, only the required information should be used to represent states and actions. Additionally, the findings demonstrated that sparse reward functions based only on the end objective may not perform as well as other reward functions that include immediate rewards. As such, it is good practice to compare several reward functions first and then choose the best one. \citet{leng2022multi} applied multi-objective RL to optimize the sequence of painting cars in a single paint shop considering the color changeover cost and adherence to sequence requirement of the assembly shop. The model was required to generate a set of Pareto optimal solutions to show the tradeoff between minimizing the changeover cost and minimizing the tardiness (i.e., adherence to the assembly shop production demand). The authors assumed a linear preference function for the objective weights in the reward function and generated an optimal policy for each of set of the weights. After training, non-dominated solutions generated by the optimal policies shaped the Pareto-frontier. Important features of these study are given in \Cref{tab:SM}. \\

\textbf{Parallel machine- } \citet{wang2020parallel} developed a scheduling model, called AlphaSchedule, to minimize tardiness in a parallel machine environment. AlphaSchedule was inspired by AlphaZero with a ResNet neural network architecture. The Proximal Policy Optimization (PPO) algorithm was employed to train the model, and integrated with Monte Carlo Tree Search (MCTS) algorithm to improve the solution quality. Numerical experiments showed that the proposed approach has a better performance and higher training efficiency than metaheuristics and simple MCTS. To address unrelated parallel machine scheduling problems, \citet{Paeng2021} proposed a DRL approach that minimizes total tardiness while considering the sequence-dependent family setup time constraint. The approach uses a novel state-action representation with dimensions invariant to production requirements and due dates. The authors also suggested a parameter sharing architecture to reduce the Deep Neural Network (DNN) parameter size. The trained DQN is able to obtain the optimal schedule in shorter time than those of the metaheuristics and simple DQN. \citet{julaiti2022stochastic} modeled a parallel machine scheduling problem with stochastic machine breakdowns and heterogeneous jobs as a partial observable MDP and solved it with a separate sampling DDPG. The proposed DDPG with separate sampling improved the estimation of state-action values and outperformed simple DDPG and heuristics in terms of optimality. Important features of these studies are also presented in \Cref{tab:SM}.\\

\begin{table}[ht]
    \caption{Summary of important features of conventional DRL models developed for single and parallel machine scheduling problems}
    \label{tab:SM}
\resizebox{\columnwidth}{!}{
\centering
\begin{tabular}{m{2cm}m{2.5cm}m{4cm}m{1.5cm}m{1.5cm}m{8cm}}

\toprule
\textbf{Authors (Year)} & \textbf{DRL method} & \textbf{Optimality Criteria} & \textbf{Agent type} & \textbf{Max. problem size (M×J)} & \textbf{Findings: \newline Optimality performance, speed, generalization}  \\       

\midrule

\multicolumn{6}{l}{\textbf{Single machine}} \\ \hline

\citet{Xie2019} & DQN & Minimize waiting times & SA & 1×dynamic job arrival & Better performance than simple heuristics.\\

\citet{leng2022multi} & DQN & Minimize tardiness and changeover cost & SA & 1×dynamic job arrival & Outperformed the metaheuristic and Q-learning in computation time, performance, convergence, and diversity of Pareto optimal solutions.\\ \\

\multicolumn{6}{l}{\textbf{Parallel machine}} \\ \hline
      
\citet{wang2020parallel} & PPO & Minimize tardiness & SA & 20×45 & Better performance and faster run-time than metaheuristics and MCTS algorithm. \\

\citet{Paeng2021} & DQN & Minimize makespan & SA & 20×100 & DQN model with DNN architecture that was featured with parameter sharing outperformed simple DQN and metaheuristics within limited time resources.\\

\citet{julaiti2022stochastic} & Separate sampling DDPG & Minimize weighted tardiness & SA & 4×100 & Better performance than simple DDPG and heuristics. \\

\bottomrule
\end{tabular}}
\end{table}

\textbf{Flow shop-} Flow Shop Scheduling Problems (FSSP) have been addressed with a variety of DRL methods including DQNs \citep{marchesano2021dynamic, kim2022reinforcement}, DQN variants \citep{yan2022deep, ren2021new, lee2022deep}, SARSA-based DRL \citep{ren2021solving}, PPO \citep{brammer2022permutation}, and Actor-Critic \citep{wagle2020use, yang2021intelligent, sun2022deep}. In \citep{marchesano2021dynamic} and \citep{kim2022reinforcement}, authors used state features represented by job characteristics of queued jobs and actions based on known dispatching rules. This method allowed for decisions to be changed dynamically in response to state changes. Another novelty improving dynamic ability was demonstrated in \citep{lee2022deep}, where the network was designed to be agnostic of changes in the number of machines. To solve the permutation Flow Shop Scheduling Problem (PFSSP), \citet{yan2022deep} and \citet{ren2021new} used DQN variants. In \citep{yan2022deep}, the DQN was augmented with a diminishing greedy rate (DQND) and the solution method also considered periodic preventative maintenance to achieve a suitable maintenance schedule. In a similar approach, \citep{ren2021new} implemented a multi-agent solution with NASH Q-Learning that uses a mean field value function to select actions rather than greedily selecting from a maximum value function.\\

In another PFSSP with machine dependent processing times, \citet{brammer2022permutation} applied the PPO algorithm in an iterative approach. The method built on a pre-trained scheduling policy and directly generated the sequencing with actions denoting which job type to be sequenced next. Similarly, an iterative Monte-Carlo-based scheduler was employed in \citep{wagle2020use}, which used schedule moments to make the model more generalizable. The scheduler allocated work pieces and processing sequences among factories using immediate rewards based on variations in the maximum completion time of each factory. For a reconfigurable flow shop where organizational structure and production set-ups could be quickly changed, \citet{yang2021intelligent} implemented an Actor-critic model to select the production mode and dispatching rules. \citet{sun2022deep} applied multi-agent DDPG for the real-time scheduling of distributed FSSP with limited buffer capacity between machines and new job insertions. Each agent controlled one of the flow shop lines and decided whether to insert newly arrived jobs on its line or adhere to its original schedule. The agents trained in centralized manner and made decisions in a decentralized manner. \Cref{tab:FSSP} summarizes important features of the studies that adopted conventional DRL models to solve FSSP.\\

\textbf{Flexible flow shop-} \citet{gil2022deep} solved a Flexible Flow Shop Scheduling Problem (FFSSP) with a single robot transfer unit using PPO. The robot had separate buffers for transferring raw, in-processing and completed materials between work stations. To keep the state and action sizes small, equivalent process machines were grouped together. The state space contained information about whether each machine group was idle or busy and whether the robot buffers were empty or full. The agent's action was determining the robot movement to a work station to which an input material should be delivered. \citet{wang2022solving} formulated job scheduling in a Resource Preemption Environment (RPE) where the operations of jobs can be processed by resources of multiple robots as a FFSSP. The RPE problem was formulated as a decentralized partially observable Markov decision process, where each job is controlled by an intelligent agent that selects an available robot to process the job based on its current partial observation. The agents were trained in a cooperative manner using the QMIX appraoch developed by \citep{rashid2018qmix}.\\

\begin{table}[ht]
    \caption{Summary of important features of conventional DRL models developed for flow shop and flexible flow shop scheduling problems}
    \label{tab:FSSP}
\resizebox{\columnwidth}{!}{
\centering
\begin{tabular}{m{2.5cm}m{2.5cm}m{4cm}m{1 cm}m{1.5cm}m{8cm}}

\toprule
\textbf{Authors (Year)} & \textbf{DRL method} & \textbf{Optimality Criteria} & \textbf{Agent type} & \textbf{Max. problem size (M×J)} & \textbf{Findings: \newline Optimality performance, speed, generalization} \\
\midrule

\multicolumn{6}{l}{\textbf{Flow shop}}  \\ \hline

\citet{wagle2020use} & Actor-critic & Maximize productivity/profit & SA & 3×dynamic job arrival  & The proposed scheduling model was applied to a simplified example plant but it could be easily scaled to more complicated industrial plants.\\

\citet{yang2021intelligent} & A2C & Minimize tardiness cost & SA & 8×150 & Outperformed meta-heuristics in solution quality and CPU times by a large margin.\\

\citet{ren2021solving} & Expected SARSA with ANN & Minimize machine idle time & SA & 5×100 & Less relative errors as compared to heuristics, SARSA, and Q-learning.\\

\citet{marchesano2021dynamic} & DQN & Maximize throughput & SA & 5×dynamic job arrival & The proposed method can be more generic by modifying some parameters.\\

\citet{brammer2022permutation} & PPO & Minimize makespan & SA & 10×10 & Superior performance to existing constructive and iterative heuristics, and exact solvers for short cutofftimes and tied with existing methods for medium and long cutofftimes.\\

\citet{kim2022reinforcement} & DQN & Minimize tardiness & SA & 1×100 & Outperformed heuristics, Q-learning, and random selection by 4\%–12\% in terms of the average tardiness and showed an average winning rate of over 77.0\%.\\

\citet{yan2022deep} & DQN with a diminishing greedy rate & Minimize makespan & SA & 20×100 & The performance is better than simple DQN and metaheuristic algorithms within limited time resources in almost all the scenarios, particularly in complicated scenarios.\\
 
\citet{ren2021new} & NASH DQN & Minimize makespan & MA  & 20×500 & Outperformed Iterative Greedy algorithm (IG) and multi-agent RL in case of large scale problems.\\

\citet{sun2022deep} & DDPG & Minimize deviation in makespan  & MA  & 3×dynamic job arrival & Better performance than multi-agent Actor-Critic, multi-agent DQN, and heuristics.\\

\multicolumn{6}{l}{\textbf{Flexible flow shop}}  \\ \hline

\citet{lee2022deep} & DQN & Minimize machine idle times & SA & 60×450 & Outperformed heuristics in terms of generating feasible solutions, maximizing throughput, maximizing demand satisfaction, and minimizing lead-time.\\

\citet{gil2022deep} & PPO & Minimize total completion time & SA & 10×800 & Outperformed DQN and heuristics.\\

\citet{wang2022solving} & QMIX & Minimize makespan & MA & 18×12 & Better performance than distributed multi-agent DRL and heuristics, generalizes, stable, and scalable.\\

\bottomrule
\end{tabular}}
\end{table}

\textbf{Job shop-} Job Shop Scheduling Problems (JSSP) have been frequently studied in the field of conventional DRL. We analyzed solving JSSP in the following subsections based on whether they employed single- or multi-agent DRL.\\

\textbf{Solving job shop scheduling problem with single agent:} The first application of DRL for machine scheduling problems (in particular JSSP) can be traced back to \citet{zhang1995high}, who applied the TD($\lambda$) algorithm to NASA’s space shuttle payload processing. The problem was to schedule a series of tasks related to each individual space shuttle mission subject to a set of temporal and resource constraints, while minimizing the makespan. The initial state of the problem was based on the critical path of the schedule, while the constraints (i.e., resource over allocations) were handled through penalties given to the agent. The approach used a time-delay neural network (TDNN) to approximate state values as a fixed feature vector for varying-length schedules. Since then, many researchers have applied DRL in various JSSP environments, most prominently with Q-learning and DQN approaches, however TD variations and policy-based methods have also been applied with success. Q-learning and TD methods for JSSPs are commonly implemented with an ANN to approximate state features and bypass the high-dimensionality dilemma of large state-space environments. The DQN methods are generally quite similar, with the substitution of more complex DNNs instead of simple ANNs. Similar to the TD($\lambda$) approach proposed by \citet{zhang1995high}, \citet{csaji2006reinforcement} implemented a TD algorithm with an ANN state estimator. Their methodology integrates the idea of market-based (negotiate-based) production control by enabling resources to bid on sequenced jobs with their respective completion cost. In this way, the agent is able to schedule resources by selecting from the possible good bidders (lowest completion cost). \citet{Riedmiller1999}, and \citet{gabel2006reducing} applied Q-Learning based solutions in similar JSSP environments consisting of three machines. Both methodologies use simple dispatching rules for two of the machines and apply their Q-Learning with ANN algorithm as an agent controlling the third machine. The latter approach \citep{gabel2006reducing} uses Monitored Q Iteration (MQI) to ensure convergence and prevent any policy degradation during advanced stages of learning. To do this, an error metric is used to continuously monitor the current policy with respect to prior experience and save the best performing policies. Early stopping was also implemented to prevent excessive simulations during learning, reaching near optimal policies while saving time and computation resources. In another Q-Learning implementation, \citet{thomas2018minerva} integrate an ANN classifier to detect bottleneck resources based on their utilization rate, waiting time, and average queue length. The classifier works interactively with the scheduling agent, periodically detecting the bottlenecks and allocating more resources to increase the overall production throughput. In many of the Q-Learning based methods, including \citet{zhang2017real}, a dummy or wait action is added to the action set which allows the agent to be idle when needed and eliminates the no-delay constraint.\\

For JSSPs with more complicated state and action spaces, basic ANNs lack the complexity required to understand such high dimensional feature sets. With the use of deeper networks, DQNs are capable of addressing such issues, and as such they have been applied in various JSSP environments for dynamic scheduling tasks \citep{Han2020, turgut2020deep, Zhou2021a, Zhao2021}. \citet{Zhang2019} chose to implement a DQN for real-time batch scheduling since DQNs can generalize well in unvisited states. Their job shop environment included many infrequent states as the scheduler not only needed to pick which batches to process, but also needed to decide if a batch should be processed immediately or wait for more jobs to join the batch. A similar problem was faced by \citet{Zhao2021} when developing a job dispatching agent in a high-dimensional, discrete state space. The authors opted to represent the discrete states as continuous values between $[0, 1]$, which improved the DQN’s convergence and versatility. Another unique approach to improve convergence was presented by \citep{lin2019smart} with the use of a multi-class DQN (MDQN). The multi-class model used a neural network with $m$ classes of neurons, one class for each of the $m$ machines. Each class contained seven neurons to represent the different dispatching rules, and their results demonstrated an improved training convergence over other methods. In a similar dispatching problem with five groups of machines, each specialized in a specific task, \citet{Altenmuller2020} use a DQN with the state space represented by machine statuses, order statuses, setup times, and buffer states. They were able to achieve success by using a composite reward function that had both local and global components. This composite reward approach was also adopted in \citep{Zhou2021a} and \citep{zhou2022reinforcement} enabling the manufacturers to change the weights of various criteria. To address delay issues, \citet{Moon2021} implemented a DQN in a cooperative multi-access edge computing ecosystem rather than using a centralized computer. Their solution employed edge devices connected to each machine that could import the pre-trained DQN model from a cloud center. The authors also implemented transfer learning to improve training convergence and generalize previous knowledge to perform multiple tasks.\\

In contrast to the value-based methods, researchers have also presented successful applications of policy-based approaches for solving JSSPs, including Trust Region Policy Optimization (TRPO) \citep{Kuhnle2019}, Proximal Policy Optimization (PPO) \citep{wang2021dynamic, Zhang2022}, and Actor-Critic \citep{Zhao2021a}. \citet{Zhang2022} demonstrated a PPO algorithm for scheduling wafer fabrication orders that uses processing times, order wait times, buffer availability and machine availability as features to represent states. Their approach implemented a reward function to optimize a weighted sum of average machine utilization and average order waiting time. One of the main advantages of PPO when compared to value-based methods is the stable learning process and fast optimization speed \citep{wang2021dynamic}. The important aspects of these single-agent studies are summarized in \Cref{tab:SAJSSP}.\\

\begin{table}[ht]
    \caption{Summary of important features of conventional single-agent DRL models developed for job shop scheduling problems}
    \label{tab:SAJSSP}
    \resizebox{\columnwidth}{!}{
        \centering
        \begin{tabular}{m{3cm}m{2.5cm}m{4cm}m{2cm}m{8.5cm}}
            \toprule
            \textbf{Authors (Year)} & \textbf{DRL method}  & \textbf{Optimality Criteria} & \textbf{Max. problem size (M$\times$J)} & \textbf{Findings: Optimality performance, speed, generalization} \\
            \midrule
            \multicolumn{5}{l}{\textbf{Job shop}} \\
            \hline
            \citet{zhang1995high} & TD($\lambda$) + TDNN & Minimize makespan & NA & Superior to non-learning solutions, and tabular TD($\lambda$) in terms of solution quality and speed. \\
            \citet{Riedmiller1999} & Q-learning + ANN & Minimize tardiness & $3\times$ dynamic job arrival & Better performance and having generalization ability compared to heuristics. \\
            \citet{csaji2006reinforcement} & TD + ANN & Minimize makespan & $20\times$ dynamic job arrival & Adaptive, when unexpected events happened, the agent should not be trained from scratch. Instead, it tried to use information from the past to speed up the computation of the new schedule. \\
            \citet{gabel2006reducing} & Modified Q-learning + ANN & Minimize tardiness & $3\times$ dynamic job arrival & Better performance than heuristics. \\
            \citet{zhang2017real} & Q-learning + ANN & Minimize cycle time & $5\times2$ & The proposed RL algorithm could generate schedules with a shorter cycle time than dispatching rules. \\
            \citet{thomas2018minerva} & Q-learning + ANN & Maximize throughput & $6\times6$ & The proposed integrated bottleneck identification and production scheduling increased the overall throughput. \\
            \citet{Zhang2019} & Q-learning + ANN & Minimize cycle time & $3\times$ dynamic job arrival & The proposed algorithm generated schedules with shorter cycle times than dispatching rules. \\
            \citep{lin2019smart} & DQN & Minimize makespan & $15\times20$ & Better performance than heuristics. \\
            \citep{Kuhnle2019} & TRPO & Maximize machine utilization & $8\times$ dynamic job arrival & Optimized both utilization of machines and lead time of orders, similar performance to heuristics. \\
            \citet{Han2020} & DDDQN & Minimize makespan & $50\times20$ & Outperforms (meta)heuristics, as fast as heuristics, faster than metaheuristics, generalizes. \\
            \citet{Altenmuller2020} & DQN & Minimize setup time violation & $10\times50$ & Time constraints with this method are managed better than heuristics. \\
            \citet{turgut2020deep} & DQN & Minimize delay time & $5\times10$ & Outperform heuristics. \\
            \citet{wang2021dynamic} & PPO & Minimize makespan & $10\times10$ & Better than heuristics, comparable to metaheuristics, as fast as heuristics (1 ms), faster than metaheuristics (28 s), generalizes. \\
            \citet{Zhao2021} & DQN & Minimize delay time & $15\times20$ & Superior performance to Q-learning, heuristics, and metaheuristics. \\
            \citet{Zhou2021a} & DQN & Maximize makespan, production cost, machine utilization & $6\times4$ & Outperforms heuristics and single-objective RL, less optimal but faster than the exact method, generalizes. \\
            \citet{Moon2021} & DQN + transfer learning & Maximize makespan & $15\times20$ & Better performance and faster convergence than the traditional DQN. \\
            \citet{Zhao2021a} & A2C & Minimize makespan and total delay & $4\times7$ & Outperformed metaheuristics and heuristics. \\
            \citet{Zhang2022} & PPO & Machine utilization and Order waiting time & $8\times$ dynamic job arrival & Outperformed TRPO and PG in optimality and speed, more flexibility in optimizing multi-objective scheduling than heuristics. \\
            \citet{zhou2022reinforcement} & DQN & Time-saving, energy-saving, machine utilization, workload distribution deviation & $6\times4$ & Better performance than CNP and single-objective RL methods, GA performed better but takes a long time to generate the schedule. \\
            \bottomrule
        \end{tabular}
    }
\end{table}

\textbf{Solving job shop scheduling problem with multi-agent DRL:} One of the first multi-agent production scheduling approaches was presented by \citet{Gabel2007} for dispatching in JSSPs. Their work demonstrated that agent-learned policies are capable of surpassing traditional dispatching policies and can generalize to other similar scheduling problems. Since then, multi-agent reinforcement learning has been applied to various JSSPs using value-based approaches such as multi-agent DQN \citep{waschneck2018deep, Dittrich2020, Baer2020, zhou2021multi} or policy-oriented methods like multi-agent Actor-Critic and DDPG \citep{liu2020actor}. Both \citet{Dittrich2020} and \citet{Baer2020} implement multi-agent DQN solutions where each order is assigned an agent to make dispatching decisions cooperatively. A centralized action-value function is trained using replay memory of collective experience from each agent. \citet{waschneck2018deep} trained DQN agents independently to control each work center. While training each agent, all other workstations were controlled by conventional dispatching rules. The final multi-agent system could produce optimal scheduling without the need of human intervention or expert knowledge. A similar approach presented by \citet{zhou2021multi} implemented a multi-agent DQN for dynamic scheduling in a JSSP, where an agent was assigned to each machine. This method implemented two separate networks, a manufacturing value network which learned from all agents to provide state values, and a scheduling policy network that specified schedules for each machine based on operation and machine states. \Cref{tab:MAJSSP} summarizes the important aspects of these reviewed studies. 

\begin{table}[ht]
    \caption{Summary of important features of conventional multi-agent DRL models developed for job shop scheduling problems}
    \label{tab:MAJSSP}
    \resizebox{\columnwidth}{!}{
        \centering
        \begin{tabular}{m{2.5cm}m{2.5cm}m{4cm}m{2cm}m{8cm}}
            \toprule
            \textbf{Authors (Year)} & \textbf{DRL method} & \textbf{Optimality Criteria} & \textbf{Max. problem size (M$\times$J)} & \textbf{Findings: Optimality performance, speed, generalization} \\
            \midrule
            \multicolumn{5}{l}{\textbf{Job shop}} \\
            \hline
            \citet{Gabel2007} & Q-learning + ANN & Minimize makespan & $15\times20$ & Its performance surpasses heuristics. \\
            \citet{waschneck2018deep} & DQN & Minimize cycle time & $7\times$ dynamic job arrival & The proposed method generated the same outcomes as heuristics without the need for any prior expert knowledge or human intervention. \\
            \citet{Dittrich2020} & DQN & Minimize tardiness & $5\times3$ & Better performance, higher flexibility, and real-time response to changes in the production system compared to capacity-based and sequential scheduling methods. \\
            \citet{Baer2020} & DQN & Minimize makespan & $6\times3$ & Generalizes, scalable, transferable. \\
            \citet{liu2020actor} & A2C & Minimize makespan & $15\times20$ & Better rescheduling performance than heuristics in a dynamic environment, similar performance to metaheuristics, exact methods, and Q-learning in a static setting, high flexibility due to short run time. \\
            \citet{zhou2021multi} & DQN & Optimize makespan and workload balance & $6\times50$ & Superior to centralized RL with DQN and CNP in optimality and speed, not better performed better than metaheuristics but works in a dynamic environment with higher speed. \\
            \bottomrule
        \end{tabular}
    }
\end{table}

\textbf{Flexible job shop- } Flexible job shop scheduling problems (FJSSPs) have been tackled with various DRL methods including both single and multi-agent approaches. Selected studies are summarized in Table \ref{tab:FJSSP}, depicting the most important aspects. Single agent methods by \citet{heger2020dynamically} and \citet{Luo2020} both implement DQNs to select dispatching rules from a predefined set. In \citep{heger2020dynamically}, the agent learned to dynamically adjust job dispatching based on four sequencing rules, while the agent in \citep{Luo2020} worked with six composite dispatching rules to allocate unprocessed operations to available machines. Implementations by \citet{lang2020integration} and \citet{Luo2021b} each apply a two-agent DQN framework to perform scheduling tasks in flexible job shop environments. \citet{lang2020integration} used one agent to select operation sequences, while the other agent allocated the jobs to machines. The authors designed their DQN agents with LSTM networks to utilize the information of earlier selection decisions that influences sequencing of future jobs. The two DQN agents presented by \citet{Luo2021b} worked in similar two-hierarchy approach where the first agent selected which optimization criteria to be used by the second agent, and the second agent would select dispatching rules accordingly. Other multi-agent scheduling approaches use many more distributed agents with single agents dedicated for each machine \citep{park2019reinforcement}, each production stage \citep{Qu2018}, or even each product \citep{Pol2021}. With an agent for each machine, \citet{Qu2018} used shared local experience to train collectively with the policy gradient method. The agents were then able to coordinate their dispatching actions to achieve optimal performance. Similarly, \citet{park2019reinforcement} developed a trained DQN that was shared amongst agents deployed at each machine. The agents determined the next setup configurations for the machine and could adapt even when the scheduling problem differed from those in the training phase. In a more complex method, \citet{liu2022deep} used the two-agent framework in combination with a distributed agent approach to address a dynamic FJSSP. In their method, two types of agents are employed in a distributed manner. Routing agents, attached to each workstation, perform machine selection to sort incoming jobs into machine queues, while sequencing agents, at each machine, decide on which jobs to process from the respective queue.

\begin{table}[ht]
    \caption{Summary of important features of conventional DRL models developed for flexible job shop scheduling problems}
    \label{tab:FJSSP}
    \resizebox{\columnwidth}{!}{
        \centering
        \begin{tabular}{m{3cm}m{1.5cm}m{4cm}m{2cm}m{2cm}m{8cm}}
            \toprule
            \textbf{Authors (Year)} & \textbf{DRL method} & \textbf{Optimality Criteria} & \textbf{Agent type} & \textbf{Max. problem size (M$\times$J)} & \textbf{Findings: Optimality performance, speed, generalization} \\
            \midrule
            \multicolumn{6}{l}{\textbf{Flexible job shop}} \\
            \hline
            \citet{Qu2018} & PG & Minimize cost & MA & $20\times10$ & Outperforms heuristics and single-agent DRL. \\
            \citet{heger2020dynamically} & DQN & Minimize tardiness & SA & $10\times$ dynamic job arrival & Outperforms heuristics, more robust than a single heuristic under unforeseen system events. \\
            \citet{park2019reinforcement} & DDQN & Minimize makespan & MA & $12\times175$ & Outperformed DQN, GA, and heuristics with 90 times less computation time over GA. \\
            \citet{lang2020integration} & DQN & Minimize makespan and tardiness & MA & $8\times20$ & Superior to metaheuristics, fast speed (0.2 s), generalizes. \\
            \citet{Luo2020} & DDQN & Minimize makespan & SA & $50\times200$ & Better performance than heuristics and Q-learning in rescheduling, generalized to untrained production configurations. \\
            \citet{Luo2021b} & DQN & Tardiness and machine utilization & Hierarchical DRL & $50\times200$ & Outperforms heuristics, Q-learning, DQN, and DDQN proposed by \citet{luo2020dynamic}, generalizes. \\
            \citet{Pol2021} & DQN & Minimizing makespan & MA & $6\times3$ & Better performance and faster than (meta)heuristics, generalizes. \\
            \citet{liu2022deep} & DDQN & Minimizing cumulative tardiness & MA & $54\times240$ & Better performance than heuristics, generalizes. \\
            \bottomrule
        \end{tabular}
    }
\end{table}

\subsection{Advanced DRL}

In this subsection, papers that adopted advanced DRL methods to solve machine scheduling problems are reviewed. As discussed in \Cref{sec:advancedDRL}, advanced DRL methods refer to the DRL models that use either encoder-decoder architectures, Graph Neural Networks (GNN), or a combination as their computational component. These studies are grouped according to their machine environment to find out how they are tailored for each specific machine environment. The important aspects of these studies are summarized in \Cref{tab:representation}.\\

\begin{table}[ht]
    \caption{Summary of advanced deep reinforcement learning methods for machine scheduling}
    \label{tab:representation}
    \resizebox{\columnwidth}{!}{%
        \centering
        \begin{tabular}{m{2cm}m{4cm}m{1.5cm}m{1.5cm}m{2cm}m{8cm}}
            \toprule
            \textbf{Authors (Year)} & \textbf{Representation} & \textbf{DRL method} & \textbf{Learning approach} & \textbf{Max. problem size (M$\times$J)} & \textbf{Performance summary: Optimality, speed, generalization} \\
            \midrule
            \multicolumn{6}{l}{\textbf{Single machine}} \\
            \hline
            \citet{Chen2019} & ENC: LSTM + DNN & A2C & L2I & 1$\times$50 & Better performance and faster (0.037 S) than Google’s OR-tools, heuristics, and conventional DRL, generalizes, scalable \\
            
            \multicolumn{6}{l}{\textbf{Parallel machine}} \\
            \hline
            \citet{Liang2022} & Enhanced PN \newline ENC: two-layer CNN \newline DEC: RNN & Not specified & L2C & 43$\times$17,670 & Superior to (meta)heuristics, and \citet{Nazari2018}, fast runtime (2.7 S), generalizes, several days of offline training \\
            
            \multicolumn{6}{l}{\textbf{Flow shop}} \\
            \hline
            \citet{wu2020real} & ENC: LSTM \newline DEC: DNN & REINFORCE & L2C & 5$\times$200 & Performance close to metaheuristics (10\% gap) but computationally faster (1.72 S) \\
            \citet{Pan2021} & ENC: RNN+FC+CNN \newline DEC: RNN + Attention & REINFORCE & L2C & 20$\times$200 & Close performance to heuristics with similar computational speed (9 MS), generalizes, 200 hours offline training \\
            \citet{cho2022minimize} & PN & PPO & L2C & 6$\times$200 & Similar performance to metaheuristics, better than heuristics, computationally faster than metaheuristics, slower than heuristics (7.9 S), generalizes, scalable \\
            \citet{dong2022minimizing} & PN \newline ENC: RNN + GIN \newline DEC: Attention mechanism & REINFORCE \newline - Self-critic & L2C & 20$\times$200 & Outperforms (meta)heuristics and PN, computationally faster (2.13 S) \\
            
            \multicolumn{6}{l}{\textbf{Flexible flow shop}} \\
            \hline
            \citet{Ni2021} & GCN & PPO & L2I & 20$\times$500 & Outperform heuristics and \citet{Li2022}, computationally faster than heuristics and slower than \citet{Li2022} (35 min), generalizes\\
            \citet{Li2022} & Bilevel \newline (DDQN + GPN) & Policy gradient for GPN & L2C + L2I & 50$\times$5000 & Outperforms metaheuristics, DRLs (DDQN, GPN, PN), and bilevel DRL (DDQN+PN) in the effectiveness and computation efficiency (170S), generalizes, scalable \\
            
            \multicolumn{6}{l}{\textbf{Job shop}} \\
            \hline
            \citet{Ren2020} & PN \newline (with glance mechanism) & A3C & L2C & 20$\times$30 & Close to meta-heuristics \\
            \citet{seito2020production} & ENC: MPNN \newline DEC: LSTM with attention & Decaying $\varepsilon$ greedy & L2C & 10$\times$50 & Outperforms heuristics, but slower (100 S), generalizes \\
            \citet{Hameed2020} & GNN & Multi-agent PPO & L2C & 4$\times$30 & 10x faster convergence \& higher stability than \citet{Dittrich2020} \\
            \citet{Zhang2020} & GIN & PPO & L2C & 20$\times$100 & Outperforms heuristics, comparable speed with heuristics (30 S), generalizes \\
            \citet{Park2021} & GNN & PPO & L2C & 20$\times$100 & Outperforms heuristics, \citet{Gabel2008}, and \citet{lin2019smart}, comparable speed to heuristics, generalizes \\
            \citet{Chen2022} & Transformer architecture & Policy gradient & L2C & 20$\times$100 & Superior to (meta)heuristics, and Google’s OR-tools in large instances in limited time (273.1 S), generalizes \\
            
            \multicolumn{6}{l}{\textbf{Flexible job shop}} \\
            \hline
            \citet{Magalhaes2021} & PN & DQN & L2I & 10$\times$10 & Comparable with the meta-heuristic, much faster than the meta-heuristic (4 S), one-day training \\
            \citet{Han2021} & Improved PN END: 1D CNN DEC: GRU RNN & REINFORCE & L2C & 15$\times$20 & Superior to heuristics, generalizes \\
            \citet{su2022self} & ENC: LSTM \newline + DNN & Q-learning \newline - Soft actor-critic & L2I & Not indicated & Outperforms metaheuristics \\
            
            \multicolumn{6}{l}{\textbf{Open shop}} \\
            \hline
            \citet{Li2020a} & ENC: GAT-DM with MHA \newline DEC: MLP & AC & L2C & 10$\times$10 & Outperform PN, less optimal but faster than Google’s OR-tools (1S) \\
            
            \bottomrule
        \end{tabular}
    }
\end{table}

\textbf{Single machine-} \citet{Chen2019} proposed a learn-to-improve framework approach for solving a single machine problem. They argued that when the problem scale becomes large, or the problem becomes complex, solution approaches that learn to construct a complete schedule from scratch lose their performance. Alternatively, their proposed approach, learns to improve an initial given solution iteratively by local adjustments until the convergence occurs. They used a Child-Sum Tree Long LSTM to encode the job features represented on a Directed Acyclic Graph (DAG). Then, the feature vector was fed to a multi-layer fully connected neural network (called region selector) to pick a region from the problem space for improvement. Another neural network (called rule selector) was trained to select a heuristic rule to change the sequence of job execution in the selected region. The region and rule selectors were trained by the Advantage Actor-Critic (A2C) algorithm, where the region selector formed the critic and the rule selector shaped the actor of the algorithm.\\

\textbf{Parallel machine-} \citet{Liang2022} used an enhanced format of pointer network to schedule laptop manufacturing in the assembly production lines of the Lenovo company. Their case study was equivalent to an unrelated parallel machine scheduling problem. The encoder was a two-layer CNN that received job and production line information as inputs. The pointer network was trained by DRL to optimize allocation of orders to production lines and sequence the processing of manufacturing orders at each production line. Despite other advanced DRL studies that simplified their scheduling problem by overlooking operational constraints, this study applied a mask mechanism to only consider allowable production time windows for each order, production capacities of each line, and job-production line compatibilities. The mask mechanism was an adjustable tensor (i.e., multidimensional matrix) in which each tensor element can be imagined as a gate to control whether it was feasible to place an order on a particular line. Also, they designed a configurable multi-objective optimization framework. They imported the weights (i.e., preferences) over the objectives as additional inputs to the encoder of the pointer network and the network was trained to generate different optimal solutions under different input weights.\\

\textbf{Flow shop-} \citet{wu2020real} formulated the problem of medical mask production as a Permutation Flow Shop Scheduling Problem (PFSSP) and solved it using a modified version of the sequence-to-sequence model. The encoder was a LSTM network, converting the input sequence of jobs to a fixed-sized feature vector. The decoder was a multi-layer fully connected neural network which generated the probability distribution for scheduling the remaining jobs at each decoding step. The encoder-decoder model was trained by the REINFORCE algorithm. Since good baseline reduces gradient variance and speeds up the learning pace of REINFORCE, Nawaz Enscore Ham heuristic and Suliman heuristic algorithms are utilized to solve each instance and provide better estimation for the baseline function of the REINFORCE algorithm. \citet{cho2022minimize} formulated the assembly stages of ship block in a production line as a PFSSP. They trained a pointer network with Proximal Policy Optimization (PPO) to determine the optimal sequence of processing orders with different characteristics and volumes.\\

\citet{Pan2021} argued that the encoder-decoder architectures are designed to solve one-dimensional combinatorial optimization problems, whereas a PFSSP is a two-dimensional problem, meaning that the size of the input vector depends on the number of jobs and machines. For this reason, they used a combination of gated recurrent units and fully connected neural networks to first aggregate the processing time of job operations and machine numbers. Then, the output was fed to a pointer network. After optimizing a problem instance by the proposed architecture, the NEH heuristic algorithm was utilized to further improve the final solution. \citet{dong2022minimizing} modified the encoder of the pointer network to solve a PFSSP considering minimization of the late work (jobs remaining after due dates arrive). They involved two encoders i.e., job and graph encoder. The job encoder consisted of an LSTM that encoded the job features. The graph encoder learnt the context information between one job and other jobs through a Graph Isomorphism Network (GIN). The jobs are represented as nodes in a complete graph and each node contained the embedded features of each job. The decoder used the output of both the graph and job encoders to determine which job to be dispatched next. The policy and baseline functions of the reinforcement learning model were trained by REINFORCE and a self-critic algorithm, respectively. During application phase, the Iterated Greedy (IG) metaheuristic was applied to the initial solutions generated by model to further improve the quality of results. Comparison results indicated that their approach outperformed the original pointer network.\\

\textbf{Flexible flow shop-} \citet{Li2022} modeled a large-scale warehouse assembly packaging line with multiple stages and multiple machines as a Flexible Flow Shop Scheduling Problem (FFSSP) and solved it using a bi-level DRL framework. The upper level employed a Double Deep Q Network (DDQN) to quickly output an initial sequence of jobs. Thereafter, the lower level received a partial sequence from the higher level to refine it and generate a high-quality solution by utilizing a Graph Pointer Network (GPN). The GPN composed of a Graph Neural Network (GNN) and a LSTM network, while the decoder operated with an attention mechanism. The GPN was trained by a policy gradient method. The higher level (DDQN) and lower level (GPN) were formulated as a leader and a follower in a bi-level game. After running the model for a certain number of iterations, the two levels ultimately converge to an equilibrium leading to an optimal sequence. \citet{Ni2021} developed a learn-to-improve framework for solving FFSSP. First, an initial feasible solution was created by the first in first out rule on a Gantt chart. Then, the problem was transferred to multiple graphs. Each machine in each work stage was represented by a directed graph in which nodes were representative of operations and directed edges showed the operation sequence. The embedding of each stage was derived using a Graph Convolutional Network (GCN) aggregated with an attention-based weighted pooling architecture. The GCN output formed the state for a PPO algorithm. Inspired by the iterated greedy algorithm, the agent's action determined the search operator and search scale. The search operator determined how jobs should be selected for reinsertion and the search scale selected what percentage of jobs should be selected.\\

\textbf{Job shop-} \citet{Ren2020} applied a pointer network to the Job Shop Scheduling Problem (JSSP). In each time step, three items, including an embedding vector of an operation of a job, the output of the previous encoding step, and the most probable operation (determined by the decoder) were fed to the encoder. The encoder generated a memory state sequence for the imported operation and fed it into the decoder. Then, the decoder outputted an operation with the highest probability in every time step considering minimization of the makespan. The Asynchronous Advantage Actor-Critic (A3C) method was utilized to find the optimal parameters of the proposed learning framework. The results showed that A3C with glimpse mechanism not only outperformed A3C without glimpse and the REINFORCE method, but also had the smallest relative error compared to the benchmark metaheuristics in small-, medium-, and large-scale instances. More recently, \citet{Chen2022} applied a transformer network architecture to the job shop scheduling problem considering makespan minimization. A disjunctive graph was used to represent the precedence constraints and machine-operation constraints of the problem. The authors used an improved version of node2vec (node to vector) algorithm \citep{grover2016node2vec} to first extract the disjunctive graph features into a sequence vector. Then, the sequence vector was fed to the transformer network. The output of the encoder was a matrix representing the encoded features of each operation. The decoder recurrently chose an operation to be arranged in each step based on the encoded matrix, the previous decoder output, and the dynamic information coming from the environment. Policy gradient was employed to train the transformer network considering the minimization of the makespan as the reward function. \citet{seito2020production} integrated a sequence-to-sequence model with a Massage Passing Neural Network (MPNN) to solve a JSSP. They mapped the JSSP problem to a Directed Acyclic Graph (DAG) in which nodes described the operations and resources, while directed edges showed the relation among the nodes. They used the MPNN to encode the graph information to a feature vector. Then, the feature vector was fed to an LSTM network with attention mechanism. The LSTM network decoded an operation to be executed at each step until all operations were scheduled. The learning framework was trained using a decaying $\varepsilon$-greedy method. Through this method, the allocation of an operation to a resource was done randomly with probability $\varepsilon$, while with probability $(1-\varepsilon)$, an allocation suggested by the trained model was chosen.\\

\citet{Hameed2020} modeled a JSSP for the case studies of robot manufacturing cells and mold injection machines as a graph where the operations and machines were given as nodes and the connections among the machines were represented as the edges. Multi-agent deep reinforcement learning was used to train a GNN. One agent was allocated to each machine node. The embedding of each machine node and the intermediate neighbor nodes were used to define the state of each agent. The action of each agent was whether or not to activate the edges connecting the nodes. Each agent was trained by PPO. \citet{Zhang2020} and \citet{Park2021} modeled the JSSP as a disjunctive graph where the static and dynamic information about operations of jobs are represented as nodes of the graph, the precedence/succeeding constraints as conjunctive edges between two nodes, and machine-sharing constraints as disjunctive edges between two nodes. \citet{Zhang2020} employed graph isomorphism network and \citet{Park2021} adopted a graph neural network to obtain the fixed dimensional embedding of the graph nodes. The GNN in both studies made their learning frameworks size-agnostic, meaning that the DRL agent was not required to be retrained for every new instance with a size larger than the training data. The learnable parameters of both GNNs shaped the policy function of the PPO algorithm. The policy function of the PPO (i.e., the actor) determined which node (operation) should be processed at each time step using the node embeddings calculated by the GNN for each operation. The critic estimated the state value function using the graph embedding calculated by the same GNN.\\

\textbf{Flexible job shop-} \citet{Magalhaes2021} used a pointer network to minimize the makespan in a dual resource constrained Flexible Job Shop Scheduling Problem (FJSSP). The encoder and decoder both consisted of Gated Recurrent Unit (GRU) neural networks. After encoding information of all operations, the decoder made one change at a time in the positional sequence of the operations until no further decrease in the makespan was achievable. The newly generated sequence was evaluated to be feasible considering the precedence constraints and machine-operation constraints. A DQN was utilized to train the learning framework. The input to the encoder and the change in the sequence of operations were assumed as the state and action of the agent, respectively. The reward function was defined to incentivize any change improving the makespan and to penalize the ones deteriorating the makespan. \citet{Han2021} employed a modified version of a pointer network to minimize the makespan of jobs in a FJSSP in a real-time fashion. A 3D disjunctive graph was used to first initialize the ready task set. Next, the dynamic and static features of the ready tasks and their corresponding machines were fed to a 1D convolutional network to obtain their graph embedding. The decoder network was modeled as a GRU with an attention mechanism to calculate the probability of each task to be operated in each time step. The convolutional layer of the encoder and the GRU of the decoder shaped the critic and actor of the DRL algorithm, respectively. The experimental results indicated that incorporating the dynamic features of the problem in addition to the static features improves the generalization capability of the learning model. \citet{su2022self} solved a FJSSP with periodic maintenance and mandatory outsourcing constraints using a self-organizing neural scheduler. The neural scheduler learnt to improve an initial solution iteratively. It consisted of an LSTM encoder, the mutation operation selector, and the mutation location selector. The problem instance was mapped to a DAG where nodes represented the descriptive data of operations and edges showed the precedence relations among operations of a job as well as the sequence of operations processed by the same machine. The encoder embedded the graph structural information into a fixed-length feature vector. Thereafter, the operation and location selectors used this feature vector as an input to their neural network to decide the best candidate operation for mutation and the new location for its placement, respectively. Both selectors were composed of multi-layer fully connected neural networks and they were trained by Q-learning and soft actor-critic, respectively.\\

\textbf{Open shop-} \citet{Li2020a} mapped the open shop scheduling problem to a DAG through representing the operations of each job as nodes and the jobs and machines as directed edges of the graph. The authors utilized a graph attention network, which consisted of an encoder and a decoder to minimize the makespan. The encoder took the processing time, positional job encoding, and positional machine encoding of each operation as the features of nodes sequentially. Then, it integrated the main information of each node and its neighbors to generate a feature vector at every encoding step using a multi-head attention mechanism and a fully connected feedforward neural network. The decoder utilized a similar structure to the encoder to output a probability for the candidate operations and the operation with the highest probability was selected to be executed in each decoding step. Once an operation was scheduled, it was masked to not be scheduled again. Since the regular attention mechanism has no memory of the past decision, a novel discount memory was introduced to the attention model that enables the previous decisions to also be considered in the current decision stage by applying a discount factor to historically acquired information. The encoder output was utilized to incrementally construct the graph structure representing the execution sequence of the operations. The learnable parameters of the graph attention network with discount memory shaped the policy network (i.e., the actor). The critic network was composed of a fully connected neural network to estimate the makespan.

\subsection{Metaheuristic-based DRL}
Since manufacturing scheduling problems are categorized as Non-deterministic Polynomial-time (NP) hard \citep{brucker1998scheduling}, numerous frameworks have been proposed in the literature using metaheuristic algorithms to find near-optimal solutions for them. The work carried out by \citet{davis1985job} who applied Genetic Algorithm (GA) to schedule a job shop problem, is one of the pioneering implementations of metaheuristic to machine scheduling problems. Since then, other metaheuristics, including ant colony optimization, grey wolf optimizer, and iterated greedy search have been widely applied to deal with different machine scheduling environments. In addition, metaheuristics have shown significant potentials and better performance as compared to exact methods and heuristics to optimize the machine schedules from a multi-objective point of view. Nonetheless, the classical metaheuristic approaches show critical limitations when dealing with dynamic and online schedule problems, inherently due to their computational complexity. To this end, and to achieve the patterns that can approximate an efficient solution, a number of studies have recently integrated metaheuristics and machine learning techniques. \citet{shahzad2016learning} have introduced a reactive tabu search to find solutions in an offline manner and generate a data set, while a decision trees classifier was employed to learn from this data and suggest similar solutions in an online setup. More recently, \citet{habib2021multiple} proposed a GA to construct a set of solutions, each of which was evaluated and scored by a simulator based on the job shop configurations and attributes. Once a decision tree was trained on this data set, it was employed as a decision-making solution in an online fashion. Nevertheless, due to the dependency of these classification and regression techniques on foreseen data and similar patterns, the development of such a scheduler was ineffective in a dynamic manufacturing system with unknown scenarios. In addition, the performance of these algorithms may drop significantly in large-scale problems.\\

To address this shortcoming, since 2020, some researchers have proposed using DRL to improve the effectiveness of the search in the problem solution space. \citet{yin2020energy} used a DRL agent to adaptively modify the population in each subgroup of a multi-objective grey wolf optimization algorithm. Kalman filtering was also added to the framework to facilitate convergence from the solution set to Pareto optimal front end. The model is proposed for a flow shop scheduling problem that takes into account two objective functions i.e., the energy saving and total makespan. \citet{chien2021agent} designed a hybrid algorithm including an $\epsilon$-greedy double deep Q network and a GA to minimize the makespan with a fast decision runtime to better cope with dynamic changes like machine shutdown, urgent jobs, and material shortage in an unrelated parallel machine scheduling setting. \citet{du2022knowledge} proposed a combination of the Estimation of Distribution Algorithm (EDA) and a Deep Q Network (DQN) to increase exploration and exploitation capabilities, respectively. EDA has excellent exploration capability, while DQN with knowledge-based decisions can improve the suggested algorithm's exploitation. At every iteration, the EDA component could investigate a bigger solution space to prevent being stuck in local optima, while the DQN component refined superior solutions for enhanced outputs. Notably, the DQN component was employed as a local search selector, which chose an appropriate local search method for specific scheduling conditions. \citet{li2022a} applied Genetic Programming (GP) method to optimize the weight of several priority dispatching rules that their combination formed high-quality composite dispatching rules. The top best composite dispatching rules generated by the last iteration of GP were selected as the action space of the DRL model. The evolution of action space through using GP effectively enhanced the quality of generated actions, hence improving the performance of the proposed method. \Cref{tab:meta} summarizes the proposed models, machine environments, objectives, comparisons and performances of the studies that proposed integrated metaheuristics and DRL in the machine scheduling domain.

\begin{table}[h]
    \caption{Summary of metaheuristic-based DRL models}
    \label{tab:meta}
\resizebox{\columnwidth}{!}{
\centering
\begin{tabular}{m{2.5cm}m{2cm}m{3cm}m{0.5cm}m{2.5cm}m{1.5cm}m{1.5cm}m{2.5cm}m{6cm}}\toprule
\textbf{Authors (Year)} &  \textbf{Proposed method} & \textbf{Machine environment} & \textbf{Multi/single objective} & \textbf{Optimality criteria} & \textbf{Realtime / offline} & \textbf{Max. problem size (M×J)} & \textbf{Benchmark} & \textbf{Performance summary} \\
\midrule
\citet{yin2020energy} & Grey Wolf+DRL & Flow shop & M & Completion time and energy consumption & Offline  & 3x3 & Simple grey wolf, particle swarm, and GA & The Kalman filter and the reinforcement learning increased the generalization and improved the grey wolf's search through the solution space.\\

\citet{chien2021agent} & HGA+DDQN & Unrelated parallel machine & S & Makespan & Offline  & 3x10   & Simple DQN and GA & Hybrid DQN-GA had better optimality performance and faster speed in real-world scenarios than other solutions.\\

\citet{du2022knowledge} & EDA+DQN & Flexible flow shop & M & Makespan and energy consumption & Offline  & 10x200 & Metaheuristics and MILP &  Outperformed not only metaheuristics but also both sim DQN as it combines the exploration ability of EDA and the exploitation ability of DQN. MILP was not able to solve problems bigger than 10 jobs x 3 machines.\\

\citet{li2022a} & GP+DQN  & Flexible job shop & M & Makespan and energy consumption & Realtime  & 15x240 & Q-Learning and DQN  &  Effectively dealt with dynamic disturbances and bridged the Sim2Real gap through continuous training of the agent.\\
\bottomrule

\end{tabular}}
\end{table}

\section{Discussion} \label{sec:discussion}
In \Cref{sec:applications}, we grouped and reviewed previous studies based on their DRL approach and machine environment. In this section, we discuss current trends, state-of-the-art and existing challenges.\\

\textbf{Publication trend- }\label{sec:trend} The number of conventional DRL models applied to machine scheduling totals 58. They have been developed since 1995. In 1995, the previous success of TD algorithm featured with a Time-Delay Neural Network (TDNN) as its value function approximator in solving the game of backgammon inspired \citet{zhang1995high} to implement this algorithm in task scheduling of NASA’s space shuttle payload processing. Since then, another four research studies utilized RL algorithms with ANN as their value function approximator by 2016. Between 2017 and 2022, there has been a surge in the number of conventional DRL models with total a of 45. This rise can be attributed to the second success of DRL models, particularly the DQN for playing Atari games presented by \citet{mnih2013playing}. Advanced DRL models have emerged since 2019 because of the remarkable performance of encoder-decoder architectures in solving other combinatorial optimization problems such as vehicle routing problems \citep{Nazari2018} and totaled 18. Lastly, metaheuristic-based DRL models have been developed since 2020 with a total number of 4.\\

\begin{figure}[h]
    \centering
    \includegraphics[width=8cm]{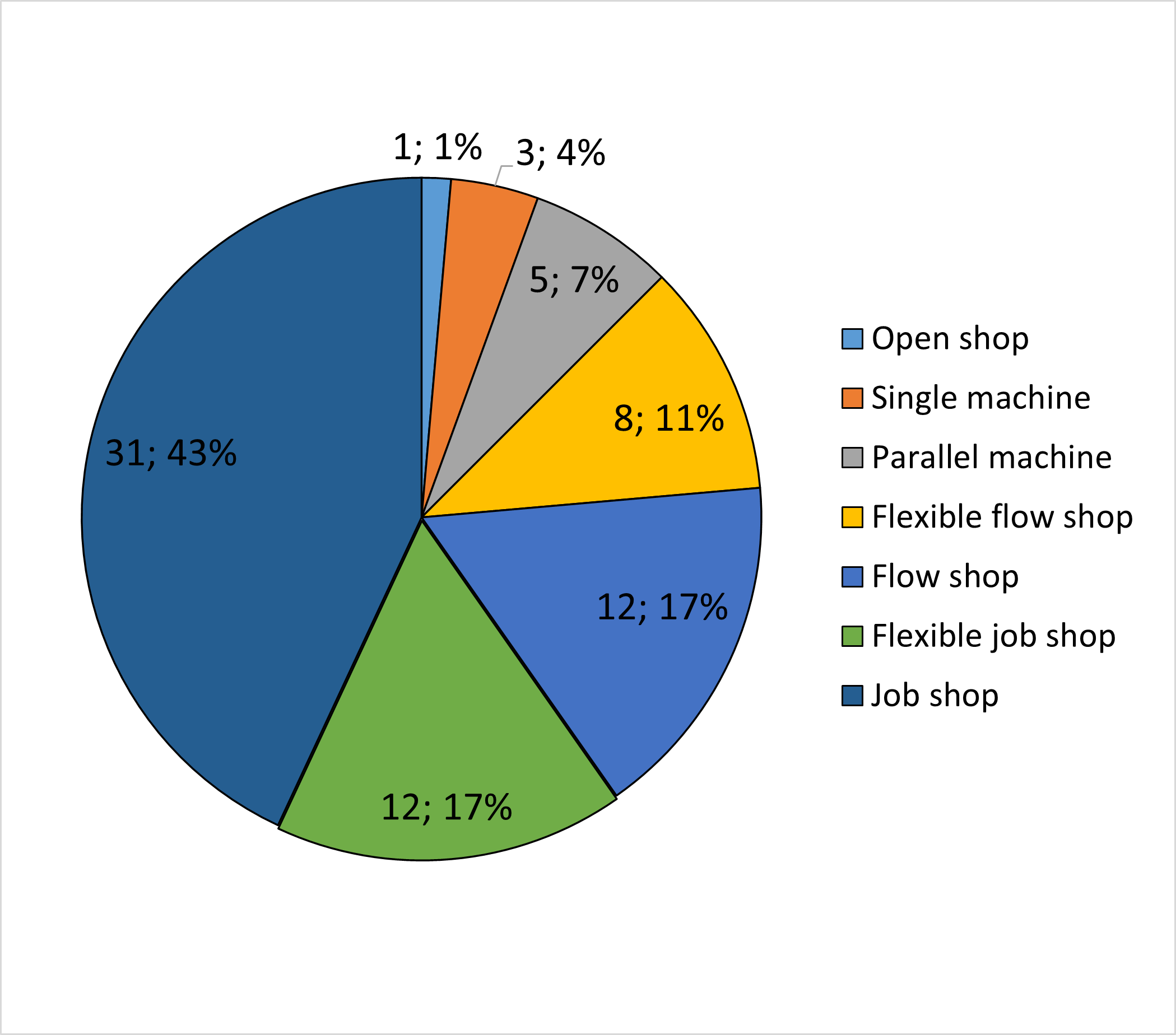}
    \caption{Distribution of publications based on their machine environment}
    \label{fig:machineEnv}
\end{figure}

\textbf{Machine environment-} Shown in \Cref{fig:machineEnv}, the majority of DRL models were developed for scheduling job shop problems, followed by flexible job shop and flow shop. Flexible flow shop ranked forth. Other machine environments including single machine, parallel machine and open shop constitute 12\% of all publications in total. The lower percentage of DRL-based papers for single and parallel machine environments is due to their less computational complexity that facilitates generating optimal solution for them using heuristics and metaheuristics. Nonetheless, according to \citep{peng2022automatic, wang2020parallel, Chen2019, Liang2022}, DRL can cope better with large-scale single and parallel machine problems compared to other methods.\\

\textbf{Job characteristics-} Previous studies mainly simplified their problem by making assumptions about their case studies. They only assumed precedence constraints, release dates, and due dates as their job characteristics and fed them to the neural network of the agent as state features. Other job characteristics, which exist in many practical cases, are considered in relatively few works. Sequence dependant setup time (sometimes referred to as changeover time) was considered as state input feature in addition to other job and machine information in \citep{Paeng2021, lee2022deep, park2021digital, heger2020dynamically, Liang2022, park2019reinforcement}. Limited buffer size for the machines was applied in \citep{Zhou2021a, Paeng2021} by considering the remaining buffer length of each machine as one of the state features. Many studies dealt with the machine breakdown by either stopping the production until the machine is fixed or rescheduling the production on other available machines reactively. \citet{yan2022deep} assumed that the deteriorating effects on machines are priory known and consequently, machine break downs were predictable. They involved a time-based maintenance policy to carry out maintenance on the machines to avoid their breakdowns. They fed the maintenance intervals as state features to the neural network of the DRL model and the agent should determine when maintenance should be carried out within a predetermined interval to have the least impact on the increase of makespan.\\ 

An important novelty in the literature to consider multiple constraints in the process of generating schedules using DRL is the mask mechanism. Different from previous works that mainly fed operational constraints as state features to the neural network, \citep{Liang2022} applied a mask mechanism to not violate the predefined production time windows, production line capacity, and machine-order compatibility. The mask mechanism was a multidimensional matrix in which each element can be imagined as a gate to control whether it was feasible to place an order on a particular line at a given time. The mask mechanism could effectively role out infeasible solutions and unlike other works, it could generalize over new instances with larger problem sizes.\\

\begin{figure}[h]
    \centering
    \includegraphics[width=8cm]{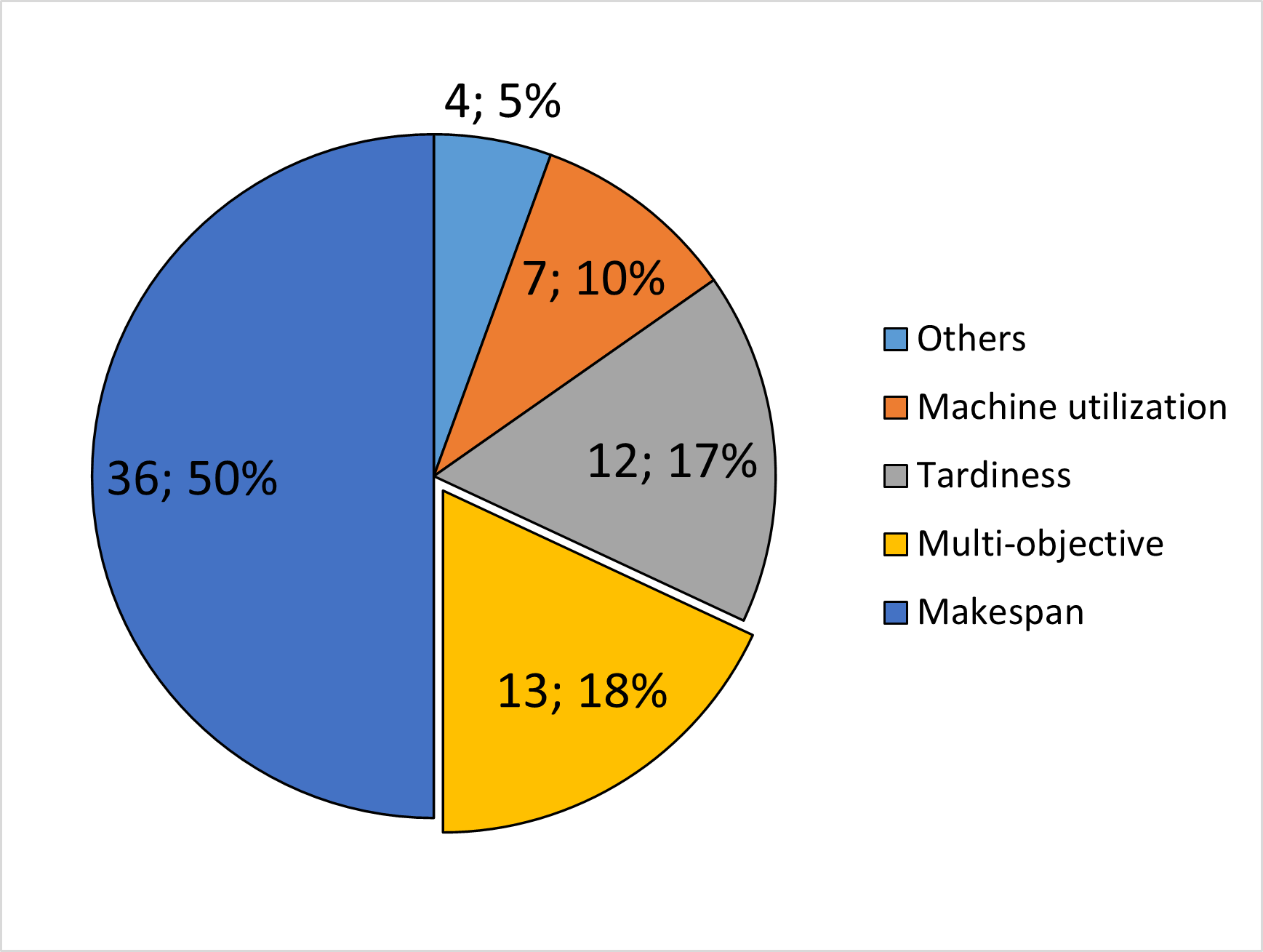}
    \caption{Distribution of publications based on their optimality criteria}
    \label{fig:opt_crt}
\end{figure}

\textbf{Optimality criteria-} \Cref{fig:opt_crt} shows the criteria that are considered as the objective function in the papers reviewed in this study to optimize the machine schedule. In single objective models, makespan is the most widely used criterion, followed by tardiness, and machine utilization. Other works with a single objective function considered minimization of either cycle time, waiting time, working-in-process cost, or setup time violations. These studies were categorized as other objectives in \Cref{fig:opt_crt} due to their small number. In total, 82\% of the works assumed a single objective for their problem and 18\% of them considered more than one objective for optimization of the schedule. They mainly applied a pre-determined weights to the objectives to calculate their weighted sum and generated a single optimal solution. As a result, no Pareto optimal set was reported and when the objective weights changed to a new set of values, the agent should be trained again. Only \citep{Liang2022} and \citep{leng2022multi} reported a Pareto set of optimal schedules. In \citep{Liang2022}, different combination of weights over the objectives were fed as inputs to the encoder of the pointer network. After training, the network learned to generate an optimal solution under each specific set of objective function weights. The authors in \citep{leng2022multi} assumed a linear preference function for the objective weights in the reward function and generated an optimal policy for each set of the weights. After training, non-dominated solutions generated by the optimal policies shaped the Pareto-frontier.\\

\textbf{Reward function design-} The reward function in each study was customized according to the single/multi-objective function that was assumed for optimizing the machine schedule. Nonetheless, the general idea is to use a combination of immediate rewards and a final reward to help the agent learn the optimal policy (e.g., see \citep{Xie2019, Ni2021, yan2022deep, Altenmuller2020, wang2021dynamic}). \citet{Xie2019} reported that only using final reward may not perform as good as other reward function schemes that include immediate rewards. Similarly, \citet{Ni2021} concluded that the evaluation of the agent performance only based on the summation of its collected immediate rewards after the end of an episode can cause learning a local optimum policy. They suggested assigning a final reward to the agent at the end of each episode would help the model to jump out of the local optimum. Another example of involving a combination of immediate and final rewards can be found in \citep{wang2021dynamic}. The ultimate goal of their DRL model was to minimize the makespan. The authors argued that the makespan cannot be obtained until all jobs are processed. Since the shorter the makespan become, the higher the machine utilization would be, they included an immediate reward based on the machine utilization level. When the end of the episode was reached and final makespan could be derived, they used the makespan value to add a final reward to the collected immediate rewards.\\

\begin{figure}[H]
    \centering
    \includegraphics[width=8cm]{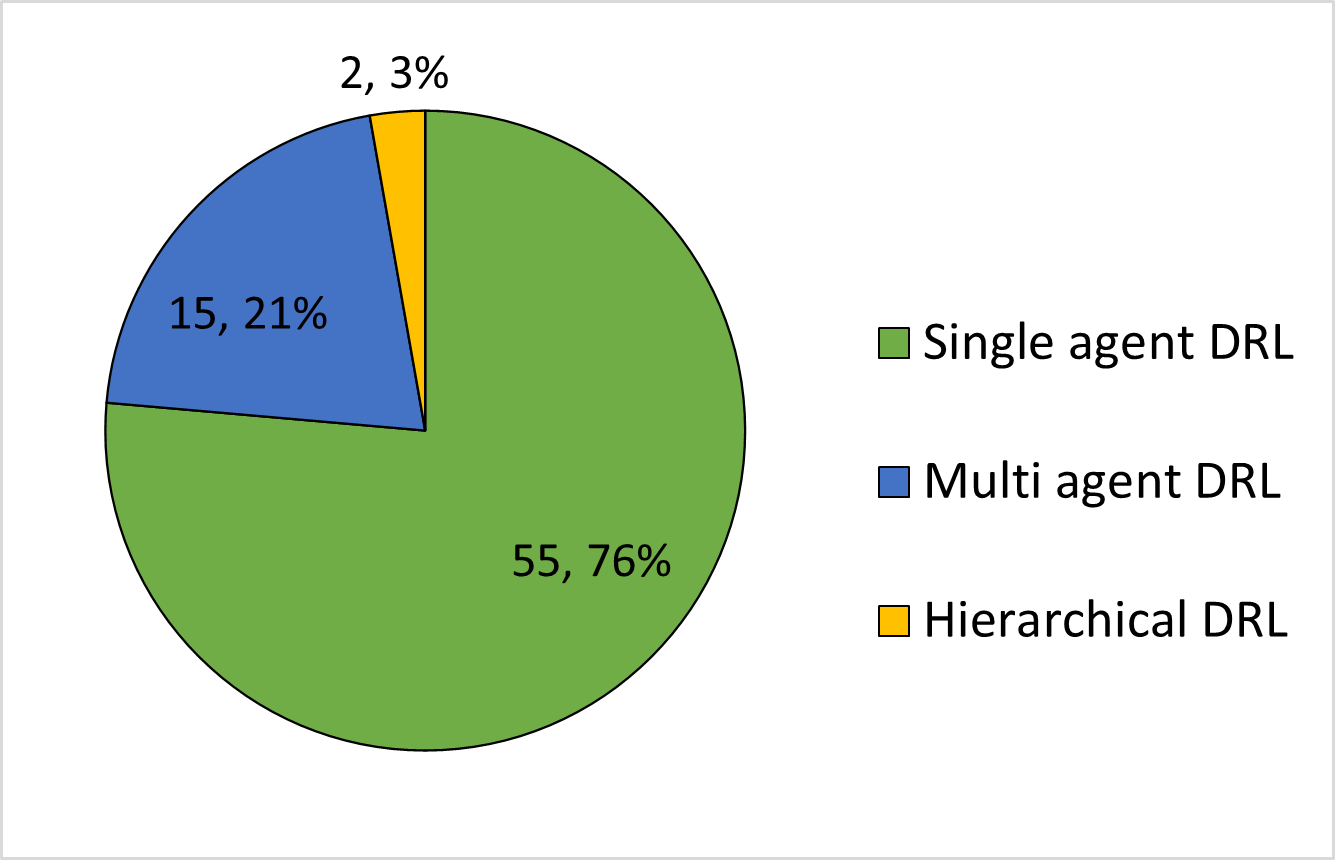}
    \caption{Distribution of publications based on their agent type}
    \label{fig:marl}
\end{figure}

\textbf{Multi-agent deep reinforcement learning-} An important aspect in the reviewed papers is whether a single-agent, multi-agent or hierarchical DRL model was adopted to solve the scheduling problem. According to \Cref{fig:marl}, majority of studies employed a single agent DRL to solve machine scheduling problems and few studies used multi-agent and hierarchical models. In single-agent DRL models, the agent received the statuses of all jobs and machines as state features and was trained to dispatch jobs to the machines. In multi-agent papers, different roles were assumed for the agents. Mostly, an agent was assigned to each machine to optimize sequence of processing waiting jobs on that machine, e.g., \citep{Gabel2007, waschneck2018deep, liu2020actor, zhou2021multi, Hameed2020, Qu2018, park2019reinforcement}. Having a model agnostic to the number of machines was the main advantages of using multi-agents since newly added machines in the environment could simply be assigned a new pretrained agent. Another way to utilize multiple agents was allocating one agent to each job \citep{Baer2020, Dittrich2020, Pol2021}. Each job was managed by a separate agent whose actions were to determine which machine from the available ones should be assigned to that job. The last approach has been applied to flexible job shop environments. This approach considers two different types of agents: (1) a routing agent that is attached to each work station and selects a machine to process the coming job, and (2) a sequencing agent that is attached to each machine and plans the optimal sequence of processing the jobs waiting in its buffer \citep{lang2020integration, liu2022deep}.\\

To train the multi-agent DRL models, different training schemes were carried out. The training approaches in the reviewed papers can be divided into decentralized learning and centralized learning. Within the first category, \citet{Gabel2007} made an optimistic assumption that all agents would have an optimal behaviour in the system and under this assumption, they can be trained independently by receiving only local information about the machine and waiting jobs that are related to them. Similarly, \citet{waschneck2018deep} trained agents in a decentralized manner. The allocated agent to each work station learned to dispatch jobs while other work stations were controlled by dispatching rules. After training all agents, they were activated to control their work centers independently. In the centralized learning category, two types of neural networks, i.e., local and global neural networks were employed. Each agent had a local network (local reply network) that contained information about local job/machine status. The agents could communicate through sending the experience of their local network to the global network (global reply memory). In this way, the global network captured the information of all agents and could provide every agent with the status of other agents \citep{liu2020actor}. According to findings of DRL models developed by \citep{Qu2018,park2019reinforcement,Dittrich2020, Baer2020, liu2020actor,zhou2021multi,liu2022deep}, centralized learning facilitated the independent and simultaneous exploration of the environment by all agents and provided the possibility for parallel training of the agents.\\

\textbf{Hierarchical deep reinforcement learning-} Shown in \Cref{fig:marl}, there are only two works in the literature that employed hierarchical DRL \citep{Luo2021b, Li2022}. \citet{Luo2021b} followed the Feudal DRL approach to reschedule production in a dynamic flexible job shop upon arrival of a new job. The higher level agent decided the temporary optimization goal for the lower level agent. The lower level agent selected a suitable dispatching rule to obtain the given goal. \citet{Li2022} solved a flexible flow shop scheduling problem using the Options framework. The upper level agent generated an initial sequence of jobs and the lower level received the given sequence to refine it and generate a high-quality solution. After running the model for a certain number of iterations, the two levels ultimately converged to an equilibrium leading to an optimal sequence. The hierarchical frameworks proposed by these studies were able to solve large scale problems and outperformed single-agent DRL methods.\\

\textbf{Optimality and computation speed-} The main goal of developing DRL models for machine scheduling is to generate near-optimal solutions in a short amount of time. Optimality is important to maintain the system's Key Performance Indicators (KPIs) at their optimal value. Fast computation speed is also important to guarantee agile response to dynamic and unpredicted changes happening in the system. The benchmark comparison of previous works indicates that DRL approaches outperformed heuristics, metaheuristics, exact methods in optimality and computation speed. Heuristics generate a feasible solution in a short computation time but their output is far from the global optima. Metaheuristics achieve near-optimal solutions but at the expense of long run time. Exact methods such as MILP guarantee optimality of the solution; however, they can only solve small-scale problems due to their practically infeasible run time in medium- and large-scale problems. The better performance of DRL approaches is due to having separate training and deployment phases. First, the agent learns an optimal policy for dispatching jobs through experiencing different instances of the scheduling problem during the training phase. Then, the agent uses the learned policy to take action in a short amount of time during the deployment phase.\\

\textbf{Generalization and scalability-} Generalization indicates the ability of the agent to prescribe an optimal schedule during the deployment phase for unseen problem instances that have the same data distribution as the training problem instances. Scalability refers to the agents' ability to optimize the large size scheduling problems. The findings of advanced DRLs in \Cref{tab:representation} show that they can better generalize their policy over previously unseen instances and are more scalable than conventional and metaheuristic-based DRLs. Their generalization ability is due to learning a parameterized policy rather than learning state values. Their scalability is comes with using encoder-decoder architectures and graph neural networks as the computational component since they are invariant to the problem size. Nonetheless, scalable and generalizable DRL models have not yet been developed for job shop, flexible job shop and open shop.\\


\textbf{Dynamic and static scheduling-}\label{sec:dayandst} According to \Cref{sec:staticdynamic}, conventional DRL models can be categorized as completely reactive dynamic scheduling tools because they are used for real-time dispatching of jobs with dynamic arrival and in case of any unpredicted events, the agent reschedules the production of waiting and semi-finished jobs based on the current state of the shop floor. Advanced DRL models are used for static and dynamic scheduling. In static settings, the agent schedules all jobs based on the available information at the beginning of scheduling horizon, whereas in dynamic settings, they cope with the scheduling problem in a predictive-reactive manner. Although conventional and advanced DRL approaches can deal with static and dynamic scheduling problems, no study considered robustness to unpredicted events and disruptions in their DRL model. Metaheuristic-based DRL models are currently limited to static scheduling. In this approach, the metaheuristic algorithm optimize the schedule, while the DRL agent optimizes the internal operation of the algorithm. Since metaheuristics require a large number of iterations to converge, they cannot respond to the dynamic changes in a timely manner at their current stage. Therefore, they are suitable for when the schedule can be generated offline.

\section{Future directions} \label{sec:future directions}
 In the previous section, we pointed out advantages and limitations of DRL models. In this section, We present the possible future directions in terms of methodology and application to address the previously discussed limitations.\\

\textbf{Scalable deep reinforcement learning-} Many industrial production systems operate in large-scale to benefit from economy of scale, and they essentially need to dynamically schedule processing of jobs on machines to ensure the production efficiency is consistently high. According to \Cref{tab:representation}, the problem sizes solved by advanced DRL models shows their ability to solve large-scale problems. However, their scalability is limited to single machine, parallel machine, and flow shop environments. In job shop, flexible job shop, and open shop, they could only solve small and medium problem sizes. Therefore, future research is needed develope scalable DRL models based on encoder-decoder architectures or graph neural networks for more complex machine environments. Another way to achieve scalable DRL models can be integration of multi-agent or hierarchical DRL models with encoder-decoders or graph neural networks. For example, the success of this integration can be found in the work conducted by \citet{Li2022}. They adopted a bi-level hierarchical DRL model where the high and low level were controlled by a DDQN network and GPN, respectively. Their framework could optimize flexible flow shop instances with 5000 jobs and 50 machines.\\

\textbf{Incorporation of manufacturing rules-} A scheduling model is implementable in practice when it considers the manufacturing rules to generate a schedule. The manufacturing rules which exist in many real-world cases consist of job-machine compatibility, manpower availability, sequence-dependant setup times, limited buffer sizes, and preemptive/non-preemptive operations. These rules were only considered by conventional and metaheuristic-based DRLs to solve small and medium-scale problems. Regarding advanced DRL which can solve large-scale problems, only \citet{Liang2022} involved some of the above manufacturing rules in their model using a novel mask mechanism to solve a parallel machine workshop. Such novelty has not been applied to other machine environments of large scale to account for complex manufacturing rules.\\

\textbf{Configurable mutli-objective scheduling optimization} In practice, a set of Key Performance Indicators (KPIs) are important to be optimized instead of a single KPI which necessitates using multi-objective optimization. Previous DRL-based scheduling models used the weighted sum of the objectives to develop the reward function and as a result, they reported a single optimal solution. Despite this, decision makers prefer to have a set of non-dominated optimal solutions rather than a single one in multi-objective problems. Among previous works, only \citep{Liang2022} generated a Pareto-optimal solution using encoder-decoder architecture for solving parallel machine scheduling. Future work can investigate generating Pareto optimal solutions using other state-of-the-art network architectures such as graph neural networks for more complex machine environments.\\

\textbf{Robustness-} Production and manufacturing systems are exposed to real-time events such as breakdowns, order cancellations, and rush orders. Successful implementation of scheduling systems in the real-world cases heavily relies on their dynamic response to these events since these events cause deviation from the original schedule. As we discussed in \Cref{sec:dayandst}, current DRL-based dynamic scheduling models can be categorized as either completely reactive or predictive-reactive.  They only focused on maximizing the efficiency when rescheduling was needed. In addition to ensuring the efficiency of the schedule, it is important to ensure the stability which attempts to minimize the deviation of the new schedule from the old schedule. A schedule that simultaneously keeps a balance between efficiency and stability is called robust according to \citep{ouelhadj2009survey}. A schedule can become robust either by predicting the uncertain events in advance and minimizing their impacts proactively by considering additional time buffers in the initial schedule (called robust proactive scheduling) or by making a balance between stability and efficiency during rescheduling (called robust predictive-reactive scheduling). Our literature review shows that no previous study addressed robustness of schedules using DRL. This accords with findings of \citep{WaubertdePuiseau2022} who reviewed reliability of DRL-based production scheduling models. Thus, one potential future research can be applying the concept of robustness to DRL-based schedulers.\\

\textbf{Integrated maintenance planing and machine scheduling-} DRL-based scheduling models in the literature mainly assumed that machines are always available during the scheduling horizon. However, this assumption might not be valid in many industrial cases. In reality, manufacturing systems stop operation of machines to apply corrective or preventive maintenance to increase machine up times and reduce maintenance costs. Lots of studies have been done to increase the accuracy of fault detection and maintenance prediction algorithms \citep{ahang2022synthesizing}. If machine scheduling and maintenance planing are treated separately, it leads to conflict between the execution of these two tasks since maintenance activities consume manufacturing time. In addition, delaying maintenance may cause increase in the probability of machine breakdown. That is why it is better to optimize both tasks simultaneously in order to improve system efficiency and system availability \citep{geurtsen2020integrated}. Only \citet{yan2022deep} employed a conventional DRL to simultaneously optimize the maintenance time and production schedules considering the deteriorating effects of machines. They assumed time-based maintenance policy and scheduling was carried out for a flow shop environment. Future works can integrate maintenance planning with production scheduling of other machine environments using other maintenance policies.\\

\textbf{Real-life implementation-} An essential part of developing scheduling optimization models is their implementation on real production and manufacturing systems. Despite that, the application of many DRL-based approaches developed in the literature was limited to demonstration in a simulation environment or on a test bench (e.g., see \citep{Zhou2021a}). Very few papers have implemented their developed DRL-based schedulers on real industrial cases. Two notable examples of real-life implementation of DRL-based scheduling are \citet{Li2022} which optimized Huawei's warehouse packaging assembly lines with multiple stages and multiple machines and \citet{Liang2022} which optimized the assembly stage of Lenovo's laptop manufacturing company. Their implementations highlighted the outstanding performance of DRL-based schedulers not only in minimizing cost and energy but also in achieving real-time optimization of the production schedule. A future research direction is applying DRL-based schedulers to other real production systems to reveal new opportunities, ideas, and areas of development.\\

\textbf{Explainable DRL-} The black box nature of DRL models causes a lack of explainability behind the DRL agent's action at each state. This property of DRL models negatively impacts the trustability of schedules generated by them. Thus, explaining the reasons behind the agent behavior is essential to draw the full potential of DRL in scheduling of real-world problems \citep{ribeiro2016should}. To this end, only \citep{Kuhnle2021} worked on the explainability of a conventional DRL approach in a small-scale job shop scheduling problem with intermediate buffer zones among machines. Future studies can investigate the application of recent advances in explainable AI to understand reasoning of agents' actions in other machine environments with more complex operational constraints and larger scales. Further, the above work provided explainability for a DRL model with a feedforward neural network, whereas there is a lack of interpretability behind the advanced DRL models with encoder-decoder architectures or graph neural networks.\\ 



\textbf{Inverse learning-} The scheduling approaches in the literature have been built based on a conventional RL concept, meaning that the reward function is predefined by a machine learning developer before training the model. However, in some cases, the reward function is not easy to define and the practitioners want the agent to learn how to schedule from the decisions made by a human expert. To address this challenge, Imitation Learning can be adopted in which the apprentice agent's purpose is to discover, via observation of demonstrations by experts, a reward function that can explain the expert's behavior \citep{piot2016bridging}. In the literature, Only \citet{ingimundardottir2018discovering} applied imitation learning to discover the priority dispatching rules in a small scale job shop environment with job release and due date as the main constraints. As the authors pointed out in \citep{ingimundardottir2018discovering}, a future work could investigate the application of imitation learning in other machine environments with different optimality criteria and job characteristics.\\

\textbf{Online learning-} Although developed DRL models are able to schedule the operations at the shop floor based on its real-time status, majority of them lack adaptability to system changes. One reason behind this limitation is offline learning. In offline learning paradigm, the agent interacts with the environment to learn the optimal policy prior to inference. During the inference phase, the model parameters remain constant and thus the agent is able to take optimal actions in a particular problem setting. Once the environment experiences variations such as addition of a new product type or a new machine, the agent is required to be trained again. One remedy to avoid retaining of the agent is adopting online learning instead of offline learning. An online algorithm keeps exploring the state-action space all the time, but at the expense of taking non-optimal actions occasionally. To control the number of non-optimal actions of the agent, recent studies have proposed the concept of conservative DRL. It adopts pretrained DRL parameters, but also keeps updating the parameter values. The reward function of conservative DRL maintains the exploration cost within a certain threshold to control the number of non-optimal actions \citep{kumar2020conservative, xu2021safely}. A future work could apply online learning methods such as conservative DRL to machine scheduling problems to address the challenges associated with offline learning.\\

\textbf{Transfer Learning-} In manufacturing plants, there can be multiple production lines with similar configuration and the ability to process specific products. Designing and maintaining a separate DRL model for each line is costly and not feasible. Transfer Learning for DRL-based scheduling models, where the knowledge learned by an agent during its training phase can be used for training another agent in another context, can solve this issue \citep{zhu2020transfer}. To this end, \citet{zheng2019manufacturing} only proposed transfer learning module for a conventional DRL approach in single machine environment to transfer dispatching policy trained on a production line of a manufacturing system to other production lines of that system. The transfer learning framework was tested on four cases, where the production line settings including job arrival speeds, urgent job probabilities (i.e., jobs with short slack time), and job processing time distributions, differ from the original case on which the agent was trained. Their proposed transfer learning module could generalize the learned dispatching rule policy to other cases than the training case and saved time and cost for data collection and model training. Future works can look at applying transfer learning to existing DRL approaches for scheduling other machine environments than single machine.\\

\textbf{Meta-deep reinforcement learning-} Meta-Deep Reinforcement Learning (Meta-DRL) is regarded as the agent capacity to understand and adapt quickly to unseen tasks using its prior experience on similar tasks. Meta-DRL can be also interpreted as multi-task learning. The core studies on this topic are related to \citet{wang2016learning} and \citet{duan2016rl}. Their proposed meta-DRL frameworks enabled their agents to learn to deal with unseen new tasks efficiently. Experiments in their papers included solving mazes with different layouts, path planning of similar robots with different physical parameters, and playing multi-armed bandit with different reward probabilities. In combinatorial optimization problems, herein machine scheduling, developing meta-DRL models to solve problem instances with the same nature but different sizes and characteristics remains as an open problem. For instance, in machine scheduling, a future direction of work would be developing a Meta-RL model that solves unseen job shop scheduling problems with different job and machine properties. One of the important outcomes of Meta-DRL models would be saving the effort and time required for designing a new DRL model for every new machine scheduling problem.

\section{Conclusions} \label{sec: conclusions}
We have comprehensively reviewed and highlighted how DRL can be used to optimize machine scheduling problems. We concluded that in the literature, the DRL approaches based on their computational component can be categorized to conventional, advanced, and metaheuristic-based DRL. While manufacturing enterprises are required to generate optimal production schedules in a short amount of time to be agile and quickly respond to the dynamic events and unpredictable changes at the shop floor, all the three DRL-based approaches have demonstrated potential to solve this problem. This is because they exhibit superior performance in terms of computation speed as well as generating near-global optimal solutions as compared to the traditional solution methods such as heuristics, and exact methods. However, this review revealed that the majority of DRL-based approaches were only implemented in simulation environments or for demonstration. This is mainly due to the fact that the agents are ad hoc i.e., they neither adapt to new problems nor are they able to optimize large-scale problems. Nonetheless, the results of previous works also indicated that multi-agent DRL and hierarchical DRL lead to scalable scheduling models whereas, advanced DRL models with encoder-decoder architectures and graph neural networks would mainly improve the generalizability. Therefore, one plausible avenue for future work would be designing a unified framework by incorporating the advantages of multi-agent/hierarchical DRL and encoder-decoders/graph neural networks to simultaneously enhance the generalization and scalability of the scheduling models. Besides, considering complex operational constraints in DRL models, configurable multi-objective optimization, explainability, and robustness are among other main research directions to be pursued in the future.

\section*{Declaration of interest}
The authors declare that they have no competing financial interests that could have influenced the work reported in this paper.

\section*{Acknowledgement}
We would like to acknowledge the financial support of NTWIST Inc. and Natural Sciences and Engineering Research Council (NSERC) Canada under the Alliance Grant ALLRP 555220 – 20, and research collaboration of NTWIST Inc. from Canada, Fraunhofer IEM, D\"{u}spohl Gmbh, and Encoway Gmbh from Germany in this research.

\printcredits
\bibliographystyle{abbrvnat}

\bibliography{main}

\end{document}